# Online Planning Algorithms for POMDPs

**Stéphane Ross**                                      STEPHANE.ROSS@MAIL.MCGILL.CA
**Joelle Pineau**                                      JPINEAU@CS.MCGILL.CA
*School of Computer Science*
*McGill University, Montreal, Canada, H3A 2A7*

**Sébastien Paquet**                                   SPAQUET@DAMAS.IFT.ULAVAL.CA
**Brahim Chaib-draa**                                  CHAIB@DAMAS.IFT.ULAVAL.CA
*Department of Computer Science and Software Engineering*
*Laval University, Quebec, Canada, G1K 7P4*

## Abstract

Partially Observable Markov Decision Processes (POMDPs) provide a rich framework for sequential decision-making under uncertainty in stochastic domains. However, solving a POMDP is often intractable except for small problems due to their complexity. Here, we focus on online approaches that alleviate the computational complexity by computing good local policies at each decision step during the execution. Online algorithms generally consist of a lookahead search to find the best action to execute at each time step in an environment. Our objectives here are to survey the various existing online POMDP methods, analyze their properties and discuss their advantages and disadvantages; and to thoroughly evaluate these online approaches in different environments under various metrics (return, error bound reduction, lower bound improvement). Our experimental results indicate that state-of-the-art online heuristic search methods can handle large POMDP domains efficiently.

## 1. Introduction

The Partially Observable Markov Decision Process (POMDP) is a general model for sequential decision problems in partially observable environments. Many planning and control problems can be modeled as POMDPs, but very few can be solved exactly because of their computational complexity: finite-horizon POMDPs are PSPACE-complete (Papadimitriou & Tsitsiklis, 1987) and infinite-horizon POMDPs are undecidable (Madani, Hanks, & Condon, 1999).

In the last few years, POMDPs have generated significant interest in the AI community and many approximation algorithms have been developed (Hauskrecht, 2000; Pineau, Gordon, & Thrun, 2003; Braziunas & Boutilier, 2004; Poupart, 2005; Smith & Simmons, 2005; Spaan & Vlassis, 2005). All these methods are *offline* algorithms, meaning that they specify, prior to the execution, the best action to execute for all possible situations. While these approximate algorithms can achieve very good performance, they often take significant time (e.g. more than an hour) to solve large problems, where there are too many possible situations to enumerate (let alone plan for). Furthermore, small changes in the environment's dynamics require recomputing the full policy, which may take hours or days.

On the other hand, online approaches (Satia & Lave, 1973; Washington, 1997; Barto, Bradtke, & Singhe, 1995; Paquet, Tobin, & Chaib-draa, 2005; McAllester & Singh, 1999; Bertsekas & Castanon, 1999; Shani, Brafman, & Shimony, 2005) try to circumvent the complexity of computing a policy by planning *online* only for the *current* information state. Online algorithms are sometimes also called agent-centered search algorithms (Koenig, 2001). Whereas an offline search would compute an exponentially large contingency plan considering all possible happenings, an online search only considers the current situation and a small horizon of contingency plans. Moreover, some of these approaches can handle environment changes without requiring more computation, which allows online approaches to be applicable in many contexts where offline approaches are not applicable, for instance, when the task to accomplish, as defined by the reward function, changes regularly in the environment. One drawback of online planning is that it generally needs to meet real-time constraints, thus greatly reducing the available planning time, compared to offline approaches.

Recent developments in online POMDP search algorithms (Paquet, Chaib-draa, & Ross, 2006; Ross & Chaib-draa, 2007; Ross, Pineau, & Chaib-draa, 2008) suggest that combining approximate offline and online solving approaches may be the most efficient way to tackle large POMDPs. In fact, we can generally compute a very rough policy offline using existing offline value iteration algorithms, and then use this approximation as a heuristic function to guide the online search algorithm. This combination enables online search algorithms to plan on shorter horizons, thereby respecting online real-time constraints and retaining a good precision. By doing an exact online search over a fixed horizon, we can guarantee a reduction in the error of the approximate offline value function. The overall time (offline and online) required to obtain a good policy can be dramatically reduced by combining both approaches.

The main purpose of this paper is to draw the attention of the AI community to online methods as a viable alternative for solving large POMDP problems. In support of this, we first survey the various existing online approaches that have been applied to POMDPs, and discuss their strengths and drawbacks. We present various combinations of online algorithms with various existing offline algorithms, such as QMDP (Littman, Cassandra, & Kaelbling, 1995), FIB (Hauskrecht, 2000), Blind (Hauskrecht, 2000; Smith & Simmons, 2005) and PBVI (Pineau et al., 2003). We then compare empirically different online approaches in two large POMDP domains according to different metrics (average discounted return, error bound reduction, lower bound improvement). We also evaluate how the available online planning time and offline planning time affect the performance of different algorithms. The results of our experiments show that many state-of-the-art online heuristic search methods are tractable in large state and observation spaces, and achieve the solution quality of state-of-the-art offline approaches at a fraction of the computational cost. The best methods achieve this by focusing the search on the most relevant future outcomes for the current decision, e.g. those that are more likely and that have high uncertainty (error) on their long-term values, such as to minimize as quickly as possible an error bound on the performance of the best action found. The tradeoff between solution quality and computing time offered by the combinations of online and offline approaches is very attractive for tackling increasingly large domains.

## 2. POMDP Model

Partially observable Markov decision processes (POMDPs) provide a general framework for acting in partially observable environments (Astrom, 1965; Smallwood & Sondik, 1973; Monahan, 1982; Kaelbling, Littman, & Cassandra, 1998). A POMDP is a generalization of the MDP model for planning under uncertainty, which gives the agent the ability to effectively estimate the outcome of its actions even when it cannot exactly observe the state of its environment.

Formally, a POMDP is represented as a tuple $(S, A, T, R, Z, O)$ where:

- $S$ is the set of all the environment states. A state is a description of the environment at a specific moment and it should capture all information relevant to the agent's decision-making process.

- $A$ is the set of all possible actions.

- $T : S \times A \times S \rightarrow [0, 1]$ is the transition function, where $T(s, a, s') = \Pr(s'|s, a)$ represents the probability of ending in state $s'$ if the agent performs action $a$ in state $s$.

- $R : S \times A \rightarrow \mathbb{R}$ is the reward function, where $R(s, a)$ is the reward obtained by executing action $a$ in state $s$.

- $Z$ is the set of all possible observations.

- $O : S \times A \times Z \rightarrow [0, 1]$ is the observation function, where $O(s', a, z) = \Pr(z|a, s')$ gives the probability of observing $z$ if action $a$ is performed and the resulting state is $s'$.

We assume in this paper that $S$, $A$ and $Z$ are all finite and that $R$ is bounded.

A key aspect of the POMDP model is the assumption that the states are not directly observable. Instead, at any given time, the agent only has access to some observation $z \in Z$ that gives incomplete information about the current state. Since the states are not observable, the agent cannot choose its actions based on the states. It has to consider a complete history of its past actions and observations to choose its current action. The history at time $t$ is defined as:

$$h_t = \{a_0, z_1, \ldots, z_{t-1}, a_{t-1}, z_t\}. \tag{1}$$

This explicit representation of the past is typically memory expensive. Instead, it is possible to summarize all relevant information from previous actions and observations in a probability distribution over the state space $S$, which is called a belief state (Astrom, 1965). A belief state at time $t$ is defined as the posterior probability distribution of being in each state, given the complete history:

$$b_t(s) = \Pr(s_t = s|h_t, b_0). \tag{2}$$

The belief state $b_t$ is a sufficient statistic for the history $h_t$ (Smallwood & Sondik, 1973), therefore the agent can choose its actions based on the current belief state $b_t$ instead of all past actions and observations. Initially, the agent starts with an initial belief state $b_0$, representing its knowledge about the starting state of the environment. Then, at any time $t$, the belief state $b_t$ can be computed from the previous belief state $b_{t-1}$, using the previous action $a_{t-1}$ and the current observation $z_t$. This is done with the belief state update function $\tau(b, a, z)$, where $b_t = \tau(b_{t-1}, a_{t-1}, z_t)$ is defined by the following equation:

$$b_t(s') = \tau(b_{t-1}, a_{t-1}, z_t)(s') = \frac{1}{\Pr(z_t | b_{t-1}, a_{t-1})} O(s', a_{t-1}, z_t) \sum_{s \in S} T(s, a_{t-1}, s') b_{t-1}(s), \quad (3)$$

where $\Pr(z|b, a)$, the probability of observing $z$ after doing action $a$ in belief $b$, acts as a normalizing constant such that $b_t$ remains a probability distribution:

$$\Pr(z|b, a) = \sum_{s' \in S} O(s', a, z) \sum_{s \in S} T(s, a, s') b(s). \quad (4)$$

Now that the agent has a way of computing its belief, the next interesting question is how to choose an action based on this belief state.

This action is determined by the agent's policy $\pi$, specifying the probability that the agent will execute any action in any given belief state, i.e. $\pi$ defines the agent's strategy for all possible situations it could encounter. This strategy should maximize the amount of reward earned over a finite or infinite time horizon. In this article, we restrict our attention to infinite-horizon POMDPs where the optimality criterion is to maximize the expected sum of discounted rewards (also called the return or discounted return). More formally, the optimal policy $\pi^*$ can be defined by the following equation:

$$\pi^* = \operatorname*{argmax}_{\pi \in \Pi} E\left[\sum_{t=0}^{\infty} \gamma^t \sum_{s \in S} b_t(s) \sum_{a \in A} R(s, a) \pi(b_t, a) \,|b_0\right], \quad (5)$$

where $\gamma \in [0, 1)$ is the discount factor and $\pi(b_t, a)$ is the probability that action $a$ will be performed in belief $b_t$, as prescribed by the policy $\pi$.

The return obtained by following a specific policy $\pi$, from a certain belief state $b$, is defined by the value function equation $V^\pi$:

$$V^\pi(b) = \sum_{a \in A} \pi(b, a) \left[R_B(b, a) + \gamma \sum_{z \in Z} \Pr(z|b, a) V^\pi(\tau(b, a, z))\right]. \quad (6)$$

Here the function $R_B(b, a)$ specifies the immediate expected reward of executing action $a$ in belief $b$ according to the reward function $R$:

$$R_B(b, a) = \sum_{s \in S} b(s) R(s, a). \quad (7)$$

The sum over $Z$ in Equation 6 is interpreted as the expected future return over the infinite horizon of executing action $a$, assuming the policy $\pi$ is followed afterwards.

Note that with the definitions of $R_B(b, a)$, $\Pr(z|b, a)$ and $\tau(b, a, z)$, one can view a POMDP as an MDP over belief states (called the belief MDP), where $\Pr(z|b, a)$ specifies the probability of moving from $b$ to $\tau(b, a, z)$ by doing action $a$, and $R_B(b, a)$ is the immediate reward obtained by doing action $a$ in $b$.

The optimal policy $\pi^*$ defined in Equation 5 represents the action-selection strategy that will maximize equation $V^\pi(b_0)$. Since there always exists a deterministic policy that maximizes $V^\pi$ for any belief states (Sondik, 1978), we will generally only consider deterministic policies (i.e. those that assign a probability of 1 to a specific action in every belief state).

The value function $V^*$ of the optimal policy $\pi^*$ is the fixed point of Bellman's equation (Bellman, 1957):

$$V^*(b) = \max_{a \in A} \left[ R_B(b, a) + \gamma \sum_{z \in Z} \Pr(z|b, a) V^*(\tau(b, a, z)) \right].$$  (8)

Another useful quantity is the value of executing a given action $a$ in a belief state $b$, which is denoted by the Q-value:

$$Q^*(b, a) = R_B(b, a) + \gamma \sum_{z \in Z} \Pr(z|b, a) V^*(\tau(b, a, z)).$$  (9)

Here the only difference with the definition of $V^*$ is that the max operator is omitted. Notice that $Q^*(b, a)$ determines the value of $a$ by assuming that the optimal policy is followed at every step after action $a$.

We now review different offline methods for solving POMDPs. These are used to guide some of the online heuristic search methods discussed later, and in some cases they form the basis of other online solutions.

## 2.1 Optimal Value Function Algorithm

One can solve optimally a POMDP for a specified finite horizon $H$ by using the value iteration algorithm (Sondik, 1971). This algorithm uses dynamic programming to compute increasingly more accurate values for each belief state $b$. The value iteration algorithm begins by evaluating the value of a belief state over the immediate horizon $t = 1$. Formally, let $V$ be a value function that takes a belief state as parameter and returns a numerical value in $\mathbb{R}$ of this belief state. The initial value function is:

$$V_1(b) = \max_{a \in A} R_B(b, a).$$  (10)

The value function at horizon $t$ is constructed from the value function at horizon $t - 1$ by using the following recursive equation:

$$V_t(b) = \max_{a \in A} \left[ R_B(b, a) + \gamma \sum_{z \in Z} \Pr(z|b, a) V_{t-1}(\tau(b, a, z)) \right].$$  (11)

The value function in Equation 11 defines the discounted sum of expected rewards that the agent can receive in the next $t$ time steps, for any belief state $b$. Therefore, the optimal policy for a finite horizon $t$ is simply to choose the action maximizing $V_t(b)$:

$$\pi_t^*(b) = \operatorname*{argmax}_{a \in A} \left[ R_B(b, a) + \gamma \sum_{z \in Z} \Pr(z|b, a) V_{t-1}(\tau(b, a, z)) \right].$$  (12)

This last equation associates an action to a specific belief state, and therefore must be computed for all possible belief states in order to define a full policy.

A key result by Smallwood and Sondik (1973) shows that the optimal value function for a finite-horizon POMDP can be represented by hyperplanes, and is therefore convex and piecewise linear. It means that the value function $V_t$ at any horizon $t$ can be represented by a set of $|S|$-dimensional hyperplanes: $\Gamma_t = \{\alpha_0, \alpha_1, \ldots, \alpha_m\}$. These hyperplanes are often called $\alpha$-vectors. Each defines a linear value function over the belief state space associated with some action $a \in A$. The value of a belief state is the maximum value returned by one of the $\alpha$-vectors for this belief state. The best action is the one associated with the $\alpha$-vector that returned the best value:

$$V_t(b) = \max_{\alpha \in \Gamma_t} \sum_{s \in S} \alpha(s) b(s). \tag{13}$$

A number of exact value function algorithms leveraging the piecewise-linear and convex aspects of the value function have been proposed in the POMDP literature (Sondik, 1971; Monahan, 1982; Littman, 1996; Cassandra, Littman, & Zhang, 1997; Zhang & Zhang, 2001). The problem with most of these exact approaches is that the number of $\alpha$-vectors needed to represent the value function grows exponentially in the number of observations at each iteration, i.e. the size of the set $\Gamma_t$ is in $O(|A||\Gamma_{t-1}|^{|Z|})$. Since each new $\alpha$-vector requires computation time in $O(|Z||S|^2)$, the resulting complexity of iteration $t$ for exact approaches is in $O(|A||Z||S|^2|\Gamma_{t-1}|^{|Z|})$. Most of the work on exact approaches has focused on finding efficient ways to prune the set $\Gamma_t$, such as to effectively reduce computation.

## 2.2 Offline Approximate Algorithms

Due to the high complexity of exact solving approaches, many researchers have worked on improving the applicability of POMDP approaches by developing approximate offline approaches that can be applied to larger problems.

In the online methods we review below, approximate offline algorithms are often used to compute lower and upper bounds on the optimal value function. These bounds are leveraged to orient the search in promising directions, to apply branch-and-bound pruning techniques, and to estimate the long term reward of belief states, as we will show in Section 3. However, we will generally want to use approximate methods which require very low computational cost. We will be particularly interested in approximations that use the underlying MDP[1] to compute lower bounds (Blind policy) and upper bounds (MDP, QMDP, FIB) on the exact value function. We also investigate the usefulness of using more precise lower bounds provided by point-based methods. We now briefly review the offline methods which will be featured in our empirical investigation. Some recent publications provide a more comprehensive overview of offline approximate algorithms (Hauskrecht, 2000; Pineau, Gordon, & Thrun, 2006).

### 2.2.1 Blind policy

A Blind policy (Hauskrecht, 2000; Smith & Simmons, 2005) is a policy where the same action is always executed, regardless of the belief state. The value function of any Blind policy is obviously a *lower bound* on $V^*$ since it corresponds to the value of one specific policy that the agent could execute in the environment. The resulting value function is specified by a set of $|A|$ $\alpha$-vectors, where each $\alpha$-vector specifies the long term expected reward of following its corresponding blind policy. These $\alpha$-vectors can be computed using a simple update rule:

$$\alpha_{t+1}^a(s) = R(s,a) + \gamma \sum_{s' \in S} T(s,a,s')\alpha_t^a(s'), \tag{14}$$

where $\alpha_0^a = \min_{s \in S} R(s,a)/(1-\gamma)$. Once these $\alpha$-vectors are computed, we use Equation 13 to obtain the lower bound on the value of a belief state. The complexity of each iteration is in $O(|A||S|^2)$, which is far less than exact methods. While this lower bound can be computed very quickly, it is usually not very tight and thus not very informative.

---

1. The MDP defined by the $(S, A, T, R)$ components of the POMDP model.

### 2.2.2 Point-Based Algorithms

To obtain tighter *lower bounds*, one can use point-based methods (Lovejoy, 1991; Hauskrecht, 2000; Pineau et al., 2003). This popular approach approximates the value function by updating it only for some selected belief states. These point-based methods sample some belief states by simulating some random interactions of the agent with the POMDP environment, and then update the value function and its gradient over those sampled beliefs. These approaches circumvent the complexity of exact approaches by sampling a small set of beliefs and maintaining at most one $\alpha$-vector per sampled belief state. Let $B$ represent the set of sampled beliefs, then the set $\Gamma_t$ of $\alpha$-vectors at time $t$ is obtained as follows:

$$
\begin{aligned}
\alpha^a(s) &= R(s,a), \\
\Gamma_t^{a,z} &= \{\alpha_i^{a,z} | \alpha_i^{a,z}(s) = \gamma \sum_{s' \in S} T(s,a,s')O(s',a,z)\alpha_i'(s'), \alpha_i' \in \Gamma_{t-1}\}, \\
\Gamma_t^b &= \{\alpha_b^a | \alpha_b^a = \alpha^a + \sum_{z \in Z} \arg\max_{\alpha \in \Gamma_t^{a,z}} \sum_{s \in S} \alpha(s)b(s), a \in A\}, \\
\Gamma_t &= \{\alpha_b | \alpha_b = \arg\max_{\alpha \in \Gamma_t^b} \sum_{s \in S} b(s)\alpha(s), b \in B\}.
\end{aligned}
\tag{15}
$$

To ensure that this gives a lower bound, $\Gamma_0$ is initialized with a single $\alpha$-vector $\alpha_0(s) = \frac{\min_{s' \in S, a \in A} R(s',a)}{1-\gamma}$. Since $|\Gamma_{t-1}| \leq |B|$, each iteration has a complexity in $O(|A||Z||S||B|(|S| + |B|))$, which is polynomial time, compared to exponential time for exact approaches.

Different algorithms have been developed using the point-based approach: PBVI (Pineau et al., 2003), Perseus (Spaan & Vlassis, 2004), HSVI (Smith & Simmons, 2004, 2005) are some of the most recent methods. These methods differ slightly in how they choose belief states and how they update the value function at these chosen belief states. The nice property of these approaches is that one can tradeoff between the complexity of the algorithm and the precision of the lower bound by increasing (or decreasing) the number of sampled belief points.

### 2.2.3 MDP

The MDP approximation consists in approximating the value function $V^*$ of the POMDP by the value function of its underlying MDP (Littman et al., 1995). This value function is an *upper bound* on the value function of the POMDP and can be computed using Bellman's equation:

$$V_{t+1}^{MDP}(s) = \max_{a \in A} \left[ R(s, a) + \gamma \sum_{s' \in S} T(s, a, s') V_t^{MDP}(s') \right].$$ (16)

The value $\hat{V}(b)$ of a belief state $b$ is then computed as $\hat{V}(b) = \sum_{s \in S} V^{MDP}(s) b(s)$. This can be computed very quickly, as each iteration of Equation 16 can be done in $O(|A||S|^2)$.

### 2.2.4 QMDP

The QMDP approximation is a slight variation of the MDP approximation (Littman et al., 1995). The main idea behind QMDP is to consider that all partial observability disappear after a single step. It assumes the MDP solution is computed to generate $V_t^{MDP}$ (Equation 16). Given this, we define:

$$Q_{t+1}^{MDP}(s, a) = R(s, a) + \gamma \sum_{s' \in S} T(s, a, s') V_t^{MDP}(s').$$ (17)

This approximation defines an $\alpha$-vector for each action, and gives an *upper bound* on $V^*$ that is tighter than $V^{MDP}$ ( i.e. $V_t^{QMDP}(b) \leq V_t^{MDP}(b)$ for all belief $b$). Again, to obtain the value of a belief state, we use Equation 13, where $\Gamma_t$ will contain one $\alpha$-vector $\alpha^a(s) = Q_t^{MDP}(s, a)$ for each $a \in A$.

### 2.2.5 FIB

The two upper bounds presented so far, QMDP and MDP, do not take into account the partial observability of the environment. In particular, *information-gathering* actions that may help identify the current state are always suboptimal according to these bounds. To address this problem, Hauskrecht (2000) proposed a new method to compute upper bounds, called the Fast Informed Bound (FIB), which is able to take into account (to some degree) the partial observability of the environment. The $\alpha$-vector update process is described as follows:

$$\alpha_{t+1}^a(s) = R(s, a) + \gamma \sum_{z \in Z} \max_{\alpha_t \in \Gamma_t} \sum_{s' \in S} O(s', a, z) T(s, a, s') \alpha_t(s').$$ (18)

The $\alpha$-vectors $\alpha_0^a$ can be initialized to the $\alpha$-vectors found by QMDP at convergence, i.e. $\alpha_0^a(s) = Q^{MDP}(s, a)$. FIB defines a single $\alpha$-vector for each action and the value of a belief state is computed according to Equation 13. FIB provides a tighter *upper bound* than QMDP ( i.e. $V_t^{FIB}(b) \leq V_t^{QMDP}(b)$ for all $b$ ). The complexity of the algorithm remains acceptable, as each iteration requires $O(|A|^2|S|^2|Z|)$ operations.

## 3. Online Algorithms for POMDPs

With offline approaches, the algorithm returns a policy defining which action to execute in every possible belief state. Such approaches tend to be applicable only when dealing with small to mid-size domains, since the policy construction step takes significant time. In large POMDPs, using a very rough value function approximation (such as the ones presented in Section 2.2) tends to substantially hinder the performance of the resulting approximate policy. Even more recent point-based methods produce solutions of limited quality in very large domains (Paquet et al., 2006).

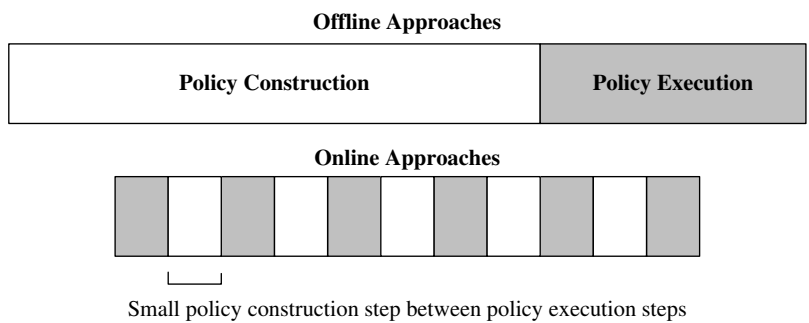

Figure 1: Comparison between offline and online approaches.

Hence in large POMDPs, a potentially better alternative is to use an online approach, which only tries to find a good local policy for the current belief state of the agent. The advantage of such an approach is that it only needs to consider belief states that are reachable from the current belief state. This focuses computation on a small set of beliefs. In addition, since online planning is done at every step (and thus generalization between beliefs is not required), it is sufficient to calculate only the *maximal value* for the current belief state, not the full optimal $\alpha$-vector. In this setting, the policy construction steps and the execution steps are interleaved with one another as shown in Figure 1. In some cases, online approaches may require a few extra execution steps (and online planning), since the policy is locally constructed and therefore not always optimal. However the policy construction time is often substantially shorter. Consequently, the *overall* time for the policy construction and execution is normally less for online approaches (Koenig, 2001). In practice, a potential limitation of online planning is when we need to meet short real-time constraints. In such case, the time available to construct the plan is very small compared to offline algorithms.

## 3.1 General Framework for Online Planning

This subsection presents a general framework for online planning algorithms in POMDPs. Subsequently, we discuss specific approaches from the literature and describe how they vary in tackling various aspects of this general framework.

An online algorithm is divided into a planning phase, and an execution phase, which are applied alternately at each time step.

In the planning phase, the algorithm is given the current belief state of the agent and computes the best action to execute in that belief. This is usually achieved in two steps. First a tree of reachable belief states from the current belief state is built by looking at several possible sequences of actions and observations that can be taken from the current belief. In this tree, the current belief is the root node and subsequent reachable beliefs (as calculated by the $\tau(b, a, z)$ function of Equation 3) are added to the tree as child nodes of their immediate previous belief. Belief nodes are represented using OR-nodes (at which we must choose an action) and actions are included in between each layer of belief nodes using AND-nodes (at which we must consider all possible observations that lead to subsequent beliefs). Then the value of the current belief is estimated by propagating value estimates up from the fringe nodes, to their ancestors, all the way to the root, according to Bellman's equation (Equation 8). The long-term value of belief nodes at the fringe is usually estimated using an approximate value function computed offline. Some methods also maintain both a lower bound and an upper bound on the value of each node. An example on how such a tree is contructed and evaluated is presented in Figure 2.

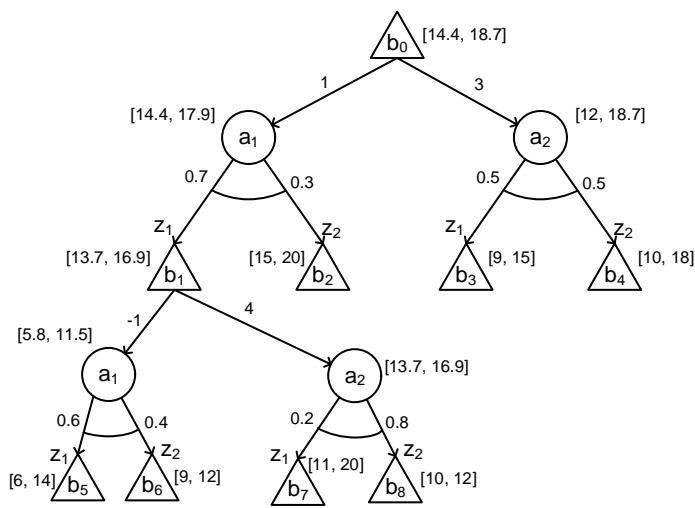

Figure 2: An AND-OR tree constructed by the search process for a POMDP with 2 actions and 2 observations. The belief states OR-nodes are represented by triangular nodes and the action AND-nodes by circular nodes. The rewards $R_B(b, a)$ are represented by values on the outgoing arcs from OR-nodes and probabilities $\Pr(z|b, a)$ are shown on the outgoing arcs from AND-nodes. The values inside brackets represent the lower and upper bounds that were computed according to Equations 19 - 22, assuming a discount factor $\gamma = 0.95$. Also notice in this example that the action $a_1$ in belief state $b_1$ could be pruned since its upper bound ($= 11.5$) is lower than the lower bound ($= 13.7$) of action $a_2$ in $b_1$.

Once the planning phase terminates, the execution phase proceeds by executing the best action found for the current belief in the environment, and updating the current belief and tree according to the observation obtained.

Notice that in general, the belief MDP could have a graph structure with cycles. Most online algorithms handle such a structure by unrolling the graph into a tree. Hence, if they reach a belief that is already elsewhere in the tree, it is duplicated. These algorithms could always be modified to handle generic graph structures by using a technique proposed in the LAO* algorithm (Hansen & Zilberstein, 2001) to handle cycles. However there are some advantages and disadvantages to doing this. A more in-depth discussion of this issue is presented in Section 5.4.

A generic online algorithm implementing the planning phase (lines 5-9) and the execution phase (lines 10-13) is presented in Algorithm 3.1. The algorithm first initializes the tree to contain only the initial belief state (line 2). Then given the current tree, the planning phase of the algorithm proceeds by first selecting the next fringe node (line 6) under which it should pursue the search (construction of the tree). The Expand function (line 7) constructs the

```
 1: Function ONLINEPOMDPSOLVER()
      Static: b_c: The current belief state of the agent.
              T: An AND-OR tree representing the current search tree.
              D: Expansion depth.
              L: A lower bound on V*.
              U: An upper bound on V*.
 2: b_c ← b_0
 3: Initialize T to contain only b_c at the root
 4: while not EXECUTIONTERMINATED() do
 5:    while not PLANNINGTERMINATED() do
 6:       b* ← CHOOSENEXTNODETOEXPAND()
 7:       EXPAND(b*, D)
 8:       UPDATEANCESTORS(b*)
 9:    end while
10:    Execute best action â for b_c
11:    Perceive a new observation z
12:    b_c ← τ(b_c, â, z)
13:    Update tree T so that b_c is the new root
14: end while
```

**Algorithm 3.1:** Generic Online Algorithm.

next reachable beliefs (using Equation 3) under the selected leaf for some pre-determined expansion depth $D$ and evaluates the approximate value function for all newly created nodes. The new approximate value of the expanded node is propagated to its ancestors via the UPDATEANCESTORS function (line 8). This planning phase is conducted until some terminating condition is met (e.g. no more planning time is available or an $\epsilon$-optimal action is found).

The execution phase of the algorithm executes the best action $\hat{a}$ found during planning (line 10) and gets a new observation from the environment (line 11). Next, the algorithm updates the current belief state and the search tree $T$ according to the most recent action $\hat{a}$ and observation $z$ (lines 12-13). Some online approaches can reuse previous computations by keeping the subtree under the new belief and resuming the search from this subtree at the next time step. In such cases, the algorithm keeps all the nodes in the tree $T$ under the new belief $b_c$ and deletes all other nodes from the tree. Then the algorithm loops back to the planning phase for the next time step, and so on until the task is terminated.

As a side note, an online planning algorithm can also be useful to improve the precision of an approximate value function computed offline. This is captured in Theorem 3.1.

**Theorem 3.1.** *(Puterman, 1994; Hauskrecht, 2000) Let $\hat{V}$ be an approximate value function and $\epsilon = \sup_b |V^*(b) - \hat{V}(b)|$. Then the approximate value $\hat{V}^D(b)$ returned by a $D$-step lookahead from belief $b$, using $\hat{V}$ to estimate fringe node values, has error bounded by $|V^*(b) - \hat{V}^D(b)| \le \gamma^D \epsilon$.*

We notice that for $\gamma \in [0, 1)$, the error converges to 0 as the depth $D$ of the search tends to $\infty$. This indicates that an online algorithm can effectively improve the performance obtained by an approximate value function computed offline, and can find an action arbitrarily close to the optimal for the current belief. However, evaluating the tree of all reachable beliefs within depth $D$ has a complexity in $O((|A||Z|)^D |S|^2)$, which is exponential in $D$. This becomes quickly intractable for large $D$. Furthermore, the planning time available during the execution may be very short and exploring all beliefs up to depth $D$ may be infeasible.

Hence this motivates the need for more efficient online algorithms that can guarantee similar or better error bounds.

To be more efficient, most of the online algorithms focus on limiting the number of reachable beliefs explored in the tree (or choose only the most relevant ones). These approaches generally differ only in how the subroutines ChooseNextNodeToExpand and Expand are implemented. We classify these approaches into three categories : Branch-and-Bound Pruning, Monte Carlo Sampling and Heuristic Search. We now present a survey of these approaches and discuss their strengths and drawbacks. A few other online algorithms do not proceed via tree search; these approaches are discussed in Section 3.5.

## 3.2 Branch-and-Bound Pruning

Branch-and-Bound pruning is a general search technique used to prune nodes that are known to be suboptimal in the search tree, thus preventing the expansion of unnecessary lower nodes. To achieve this in the AND-OR tree, a lower bound and an upper bound are maintained on the value $Q^*(b, a)$ of each action $a$, for every belief $b$ in the tree. These bounds are computed by first evaluating the lower and upper bound for the fringe nodes of the tree. These bounds are then propagated to parent nodes according to the following equations:

$$L_T(b) = \begin{cases} L(b), & \text{if } b \in \mathcal{F}(T) \\ \max_{a \in A} L_T(b, a), & \text{otherwise} \end{cases} \quad (19)$$

$$L_T(b, a) = R_B(b, a) + \gamma \sum_{z \in Z} \Pr(z|b, a) L_T(\tau(b, a, z)), \quad (20)$$

$$U_T(b) = \begin{cases} U(b), & \text{if } b \in \mathcal{F}(T) \\ \max_{a \in A} U_T(b, a), & \text{otherwise} \end{cases} \quad (21)$$

$$U_T(b, a) = R_B(b, a) + \gamma \sum_{z \in Z} \Pr(z|b, a) U_T(\tau(b, a, z)), \quad (22)$$

where $\mathcal{F}(T)$ denotes the set of fringe nodes in tree $T$, $U_T(b)$ and $L_T(b)$ represent the upper and lower bounds on $V^*(b)$ associated to belief state $b$ in the tree $T$, $U_T(b, a)$ and $L_T(b, a)$ represent corresponding bounds on $Q^*(b, a)$, and $L(b)$ and $U(b)$ are the bounds used at fringe nodes, typically computed offline. These equations are equivalent to Bellman's equation (Equation 8), however they use the lower and upper bounds of the children, instead of $V^*$. Several techniques presented in Section 2.2 can be used to quickly compute lower bounds (Blind policy) and upper bounds (MDP, QMDP, FIB) offline.

Given these bounds, the idea behind Branch-and-Bound pruning is relatively simple: if a given action $a$ in a belief $b$ has an upper bound $U_T(b, a)$ that is lower than another action $\tilde{a}$'s lower bound $L_T(b, \tilde{a})$, then we know that $\tilde{a}$ is guaranteed to have a value $Q^*(b, \tilde{a}) \geq Q^*(b, a)$. Thus $a$ is suboptimal in belief $b$. Hence that branch can be pruned and no belief reached by taking action $a$ in $b$ will be considered.

### 3.2.1 RTBSS

The Real-Time Belief Space Search (RTBSS) algorithm uses a Branch-and-Bound approach to compute the best action to take in the current belief (Paquet et al., 2005, 2006). Starting

---

1: **Function** EXPAND($b$, $d$)
   **Inputs:** $b$: The belief node we want to expand.
          $d$: The depth of expansion under $b$.
   **Static:** $T$: An AND-OR tree representing the current search tree.
          $L$: A lower bound on $V^*$.
          $U$: An upper bound on $V^*$.

2: **if** $d = 0$ **then**
3:    $L_T(b) \leftarrow L(b)$
4: **else**
5:    Sort actions $\{a_1, a_2, \ldots, a_{|A|}\}$ such that $U(b, a_i) \geq U(b, a_j)$ if $i \leq j$
6:    $i \leftarrow 1$
7:    $L_T(b) \leftarrow -\infty$
8:    **while** $i \leq |A|$ **and** $U(b, a_i) > L_T(b)$ **do**
9:       $L_T(b, a_i) \leftarrow R_B(b, a_i) + \gamma \sum_{z \in Z} \Pr(z|b, a_i) \text{EXPAND}(\tau(b, a_i, z), d-1)$
10:      $L_T(b) \leftarrow \max\{L_T(b), L_T(b, a_i)\}$
11:      $i \leftarrow i + 1$
12:   **end while**
13: **end if**
14: **return** $L_T(b)$

**Algorithm 3.2:** Expand subroutine of RTBSS.

---

from the current belief, it expands the AND-OR tree in a depth-first search fashion, up to some pre-determined search depth $D$. The leaves of the tree are evaluated by using a lower bound computed offline, which is propagated upwards such that a lower bound is maintained for each node in the tree.

To limit the number of nodes explored, Branch-and-Bound pruning is used along the way to prune actions that are known to be suboptimal, thus excluding unnecessary nodes under these actions. To maximize pruning, RTBSS expands the actions in *descending order* of their upper bound (first action expanded is the one with highest upper bound). By expanding the actions in this order, one never expands an action that could have been pruned if actions had been expanded in a different order. Intuitively, if an action has a higher upper bound than the other actions, then it cannot be pruned by the other actions since their lower bound will never exceed their upper bound. Another advantage of expanding actions in descending order of their upper bound is that as soon as we find an action that can be pruned, then we also know that all remaining actions can be pruned, since their upper bounds are necessarily lower. The fact that RTBSS proceeds via a depth-first search also increases the number of actions that can be pruned since the bounds on expanded actions become more precise due to the search depth.

In terms of the framework in Algorithm 3.1, RTBSS requires the CHOOSENEXTNODETO-EXPAND subroutine to simply return the current belief $b_c$. The UPDATEANCESTORS function does not need to perform any operation since $b_c$ has no ancestor (root of the tree $T$). The EXPAND subroutine proceeds via depth-first search up to a fixed depth $D$, using Branch-and-Bound pruning, as mentioned above. This subroutine is detailed in Algorithm 3.2. After this expansion is performed, PLANNINGTERMINATED evaluates to true and the best action found is executed. At the end of each time step, the tree $T$ is simply reinitialized to contain the new current belief at the root node.

The efficiency of RTBSS depends largely on the precision of the lower and upper bounds computed offline. When the bounds are tight, more pruning will be possible, and the search will be more efficient. If the algorithm is unable to prune many actions, searching will be limited to short horizons in order to meet real-time constraints. Another drawback of RTBSS is that it explores all observations equally. This is inefficient since the algorithm could explore parts of the tree that have a small probability of occurring and thus have a small effect on the value function. As a result, when the number of observations is large, the algorithm is limited to exploring over a short horizon.

As a final note, since RTBSS explores all reacheable beliefs within depth $D$ (except some reached by suboptimal actions), then it can guarantee the same error bound as a $D$-step lookahead (see Theorem 3.1). Therefore, the online search directly improves the precision of the original (offline) value bounds by a factor $\gamma^D$. This aspect was confirmed empirically in different domains where the RTBSS authors combined their online search with bounds given by various offline algorithms. In some cases, their results showed a tremendous improvement of the policy given by the offline algorithm (Paquet et al., 2006).

## 3.3 Monte Carlo Sampling

As mentioned above, expanding the search tree fully over a large set of observations is infeasible except for shallow depths. In such cases, a better alternative may be to sample a subset of observations at each expansion and only consider beliefs reached by these sampled observations. This reduces the branching factor of the search and allows for deeper search within a set planning time. This is the strategy employed by Monte Carlo algorithms.

### 3.3.1 McAllester and Singh

The approach presented by McAllester and Singh (1999) is an adaptation of the online MDP algorithm presented by Kearns, Mansour, and Ng (1999). It consists of a depth-limited search of the AND-OR tree up to a certain fixed horizon $D$ where instead of exploring all observations at each action choice, $C$ observations are sampled from a generative model. The probabilities $\Pr(z|b,a)$ are then approximated using the observed frequencies in the sample. The advantage of such an approach is that sampling an observation from the distribution $\Pr(z|b,a)$ can be achieved very efficiently in $O(\log|S| + \log|Z|)$, while computing the exact probabilities $\Pr(z|b,a)$ is in $O(|S|^2)$ for each observation $z$. Thus sampling can be useful to alleviate the complexity of computing $\Pr(z|b,a)$, at the expense of a less precise estimate. Nevertheless, a few samples is often sufficient to obtain a good estimate as the observations that have the most effect on $Q^*(b,a)$ (i.e. those which occur with high probability) are more likely to be sampled. The authors also apply belief state factorization as in Boyen and Koller (1998) to simplify the belief state calculations.

For the implementation of this algorithm, the Expand subroutine expands the tree up to fixed depth $D$, using Monte Carlo sampling of observations, as mentioned above (see Algorithm 3.3). At the end of each time step, the tree $T$ is reinitialized to contain only the new current belief at the root.

Kearns et al. (1999) derive bounds on the depth $D$ and the number of samples $C$ needed to obtain an $\epsilon$-optimal policy with high probability and show that the number of samples required grows exponentially in the desired accuracy. In practice, the number of samples required is infeasible given realistic online time constraints. However, performance in terms of returns is usually good even with many fewer samples.

```
 1: Function EXPAND(b, d)
    Inputs:  b: The belief node we want to expand.
             d: The depth of expansion under b.
    Static:  T: An AND-OR tree representing the current search tree.
             C: The number of observations to sample.
 2: if d = 0 then
 3:     L_T(b) ← max_{a∈A} R_B(b, a)
 4: else
 5:     L_T(b) ← −∞
 6:     for all  a ∈ A  do
 7:         Sample Z = {z_1, z_2, . . . z_C} from distribution Pr(z|b, a)
 8:         L_T(b, a) ← R_B(b, a) + γ Σ_{z∈Z|N_z(Z)>0} (N_z(Z)/C) EXPAND(τ̂(b, a, z), d − 1)
 9:         L_T(b) ← max{L_T(b), L_T(b, a)}
10:     end for
11: end if
12: return L_T(b)
```

**Algorithm 3.3:** EXPAND subroutine of McAllester and Singh's Algorithm.

One inconvenience of this method is that no action pruning can be done since Monte Carlo estimation is not guaranteed to correctly propagate the lower (and upper) bound property up the tree. In their article, the authors simply approximate the value of the fringe belief states by the immediate reward $R_B(b, a)$; this could be improved by using any good estimate of $V^*$ computed offline. Note also that this approach may be difficult to apply in domains where the number of actions $|A|$ is large. Of course this may further impact performance.

### 3.3.2 ROLLOUT

Another similar online Monte Carlo approach is the Rollout algorithm (Bertsekas & Castanon, 1999). The algorithm requires an initial policy (possibly computed offline). At each time step, it estimates the future expected value of each action, assuming the initial policy is followed at future time steps, and executes the action with highest estimated value. These estimates are obtained by computing the average discounted return obtained over a set of $M$ sampled trajectories of depth $D$. These trajectories are generated by first taking the action to be evaluated, and then following the initial policy in subsequent belief states, assuming the observations are sampled from a generative model. Since this approach only needs to consider different actions at the root belief node, the number of actions $|A|$ only influences the branching factor at the first level of the tree. Consequently, it is generally more scalable than McAllester and Singh's approach. Bertsekas and Castanon (1999) also show that with enough sampling, the resulting policy is guaranteed to perform at least as well as the initial policy with high probability. However, it generally requires many sampled trajectories to provide substantial improvement over the initial policy. Furthermore, the initial policy has a significant impact on the performance of this approach. In particular, in some cases it might be impossible to improve the return of the initial policy by just changing the immediate action (e.g. if several steps need to be changed to reach a specific subgoal to which higher rewards are associated). In those cases, the Rollout policy will never improve over the initial policy.

```
 1: Function Expand(b, d)
    Inputs:  b: The belief node we want to expand.
             d: The depth of expansion under b.
    Static:  T: An AND-OR tree representing the current search tree.
             Π: A set of initial policies.
             M: The number of trajectories of depth d to sample.

 2: L_T(b) ← −∞
 3: for all  a ∈ A  do
 4:    for all  π ∈ Π  do
 5:       Q̂^π(b, a) ← 0
 6:       for  i = 1 to M  do
 7:          b̃ ← b
 8:          ã ← a
 9:          for  j = 0 to d  do
10:             Q̂^π(b, a) ← Q̂^π(b, a) + (1/M) γ^j R_B(b̃, ã)
11:             z ← SampleObservation(b̃, ã)
12:             b̃ ← τ(b̃, ã, z)
13:             ã ← π(b̃)
14:          end for
15:       end for
16:    end for
17:    L_T(b, a) = max_{π ∈ Π} Q̂^π(b, a)
18: end for
```

**Algorithm 3.4:** Expand subroutine of the Parallel Rollout Algorithm.

To address this issue relative to the initial policy, Chang, Givan, and Chong (2004) introduced a modified version of the algorithm, called Parallel Rollout. In this case, the algorithm starts with a *set* of initial policies. Then the algorithm proceeds as Rollout for each of the initial policies in the set. The value considered for the immediate action is the maximum over that set of initial policies, and the action with highest value is executed. In this algorithm, the policy obtained is guaranteed to perform at least as well as the best initial policy with high probability, given enough samples. Parallel Rollout can handle domains with a large number of actions and observations, and will perform well when the set of initial policies contain policies that are good in different regions of the belief space.

The Expand subroutine of the Parallel Rollout algorithm is presented in Algorithm 3.4. The original Rollout algorithm by Bertsekas and Castanon (1999) is the same algorithm in the special case where the set of initial policies Π contains only one policy. The other subroutines proceed as in McAllester and Singh's algorithm.

## 3.4 Heuristic Search

Instead of using Branch-and-Bound pruning or Monte Carlo sampling to reduce the branching factor of the search, heuristic search algorithms try to focus the search on the most relevant reachable beliefs by using heuristics to select the best fringe beliefs node to expand. The most relevant reachable beliefs are the ones that would allow the search algorithm to make good decisions as quickly as possible, i.e. by expanding as few nodes as possible.

There are three different online heuristic search algorithms for POMDPs that have been proposed in the past: Satia and Lave (1973), BI-POMDP (Washington, 1997) and AEMS (Ross & Chaib-draa, 2007). These algorithms all maintain both lower and upper bounds on the value of each node in the tree (using Equations 19 - 22) and only differ in the specific heuristic used to choose the next fringe node to expand in the AND/OR tree. We first present the common subroutines for these algorithms, and then discuss their different heuristics.

Recalling the general framework of Algorithm 3.1, three steps are interleaved several times in heuristic search algorithms. First, the best fringe node to expand (according to the heuristic) in the current search tree $T$ is found. Then the tree is expanded under this node (usually for only one level). Finally, ancestor nodes' values are updated; their values must be updated before we choose the next node to expand, since the heuristic value usually depends on them. In general, heuristic search algorithms are slightly more computationally expensive than standard depth- or breadth-first search algorithms, due to the extra computations needed to select the best fringe node to expand, and the need to update ancestors at each iteration. This was not required by the previous methods using Branch-and-Bound pruning and/or Monte Carlo sampling. If the complexity of these extra steps is too high, then the benefit of expanding only the most relevant nodes might be outweighed by the lower number of nodes expanded (assuming a fixed planning time).

In heuristic search algorithms, a particular heuristic value is associated with every fringe node in the tree. This value should indicate how important it is to expand this node in order to improve the current solution. At each iteration of the algorithm, the goal is to find the fringe node that maximizes this heuristic value among all fringe nodes. This can be achieved efficiently by storing in each node of the tree a reference to the best fringe node to expand within its subtree, as well as its associated heuristic value. In particular, the root node always contains a reference to the best fringe node for the whole tree. When a node is expanded, its ancestors are the only nodes in the tree where the best fringe node reference, and corresponding heuristic value, need to be updated. These can be updated efficiently by using the references and heuristic values stored in the lower nodes via a dynamic programming algorithm, described formally in Equations 23 and 24. Here $H_T^*(b)$ denotes the highest heuristic value among the fringe nodes in the subtree of $b$, $b_T^*(b)$ is a reference to this fringe node, $H_T(b)$ is the basic heuristic value associated to fringe node $b$, and $H_T(b, a)$ and $H_T(b, a, z)$ are factors that weigh this basic heuristic value at each level of the tree $T$. For example, $H_T(b, a, z)$ could be $\Pr(z|b, a)$ in order to give higher weight to (and hence favor) fringe nodes that are reached by the most likely observations.

$$
\begin{aligned}
H_T^*(b) &= \begin{cases} H_T(b) & \text{if } b \in \mathcal{F}(T) \\ \max_{a \in A} H_T(b, a) H_T^*(b, a) & \text{otherwise} \end{cases} \\
H_T^*(b, a) &= \max_{z \in Z} H_T(b, a, z) H_T^*(\tau(b, a, z))
\end{aligned}
\tag{23}
$$

$$
\begin{aligned}
b_T^*(b) &= \begin{cases} b & \text{if } b \in \mathcal{F}(T) \\ b_T^*(b, a_b^T) & \text{otherwise} \end{cases} \\
b_T^*(b, a) &= b_T^*(\tau(b, a, z_{b,a}^T)) \\
a_b^T &= \operatorname{argmax}_{a \in A} H_T(b, a) H_T^*(b, a) \\
z_{b,a}^T &= \operatorname{argmax}_{z \in Z} H_T(b, a, z) H_T^*(\tau(b, a, z))
\end{aligned}
\tag{24}
$$

This procedure finds the fringe node $b \in \mathcal{F}(T)$ that maximizes the overall heuristic value $H_T(b_c, b) = H_T(b) \prod_{i=1}^{d_T(b)} H_T(b_i, a_i) H_T(b_i, a_i, z_i)$, where $b_i$, $a_i$ and $z_i$ represent the $i^{th}$ belief, action and observation on the path from $b_c$ to $b$ in $T$, and $d_T(b)$ is the depth of fringe node $b$. Note that $H_T^*$ and $b_T^*$ are only updated in the ancestor nodes of the last expanded node. By reusing previously computed values for the other nodes, we have a procedure

---

1: **Function** Expand($b$)
   **Inputs:** $b$: An OR-Node we want to expand.
   **Static:** $b_c$: The current belief state of the agent.
   $\quad\quad\quad$ $T$: An AND-OR tree representing the current search tree.
   $\quad\quad\quad$ $L$: A lower bound on $V^*$.
   $\quad\quad\quad$ $U$: An upper bound on $V^*$.

2: **for each** $a \in A$ **do**
3: $\quad$ **for each** $z \in Z$ **do**
4: $\quad\quad$ $b' \leftarrow \tau(b, a, z)$
5: $\quad\quad$ $U_T(b') \leftarrow U(b')$
6: $\quad\quad$ $L_T(b') \leftarrow L(b')$
7: $\quad\quad$ $H_T^*(b') \leftarrow H_T(b')$
8: $\quad\quad$ $b_T^*(b') \leftarrow b'$
9: $\quad$ **end for**
10: $\quad$ $L_T(b,a) \leftarrow R_B(b,a) + \gamma \sum_{z \in Z} \Pr(z|b,a) L_T(\tau(b,a,z))$
11: $\quad$ $U_T(b,a) \leftarrow R_B(b,a) + \gamma \sum_{z \in Z} \Pr(z|b,a) U_T(\tau(b,a,z))$
12: $\quad$ $z_{b,a}^T \leftarrow \operatorname{argmax}_{z \in Z} H_T(b,a,z) H_T^*(\tau(b,a,z))$
13: $\quad$ $H_T^*(b,a) = H_T(b,a,z_{b,a}^T) H_T^*(\tau(b,a,z_{b,a}^T))$
14: $\quad$ $b_T^*(b,a) \leftarrow b_T^*(\tau(b,a,z_{b,a}^T))$
15: **end for**
16: $L_T(b) \leftarrow \max\{\max_{a \in A} L_T(b,a), L_T(b)\}$
17: $U_T(b) \leftarrow \min\{\max_{a \in A} U_T(b,a), U_T(b)\}$
18: $a_b^T \leftarrow \operatorname{argmax}_{a \in A} H_T(b,a) H_T^*(b,a)$
19: $H_T^*(b) \leftarrow H_T(b, a_b^T) H_T^*(b, a_b^T)$
20: $b_T^*(b) \leftarrow b_T^*(b, a_b^T)$

---

**Algorithm 3.5:** Expand : Expand subroutine for heuristic search algorithms.

that can find the best fringe node to expand in the tree in time linear in the depth of the tree (versus exponential in the depth of the tree for the exhaustive search through all fringe nodes). These updates are performed in both the Expand and the UpdateAncestors subroutines, each of which is described in more detail below. After each iteration, the ChooseNextNodeToExpand subroutine simply returns the reference to this best fringe node stored in the root of the tree, i.e. $b_T^*(b_c)$.

The Expand subroutine used by heuristic search methods is presented in Algorithm 3.5. It performs a one-step lookahead under the fringe node $b$. The main difference with respect to previous methods in Sections 3.2 and 3.3 is that the heuristic value and best fringe node to expand for these new nodes are computed at lines 7-8 and 12-14. The best leaf node in $b$'s subtree and its heuristic value are then computed according to Equations 23 and 24 (lines 18-20).

The UpdateAncestors function is presented in Algorithm 3.6. The goal of this function is to update the bounds of the ancestor nodes, and find the best fringe node to expand next. Starting from a given OR-Node $b'$, the function simply updates recursively the ancestor nodes of $b'$ in a bottom-up fashion, using Equations 19-22 to update the bounds and Equations 23-24 to update the reference to the best fringe to expand and its heuristic value. Notice that the UpdateAncestors function can reuse information already stored in the node objects, such that it does not need to recompute $\tau(b,a,z)$, $\Pr(z|b,a)$ and $R_B(b,a)$. However it may need to recompute $H_T(b,a,z)$ and $H_T(b,a)$ according to the new bounds, depending on how the heuristic is defined.

Due to the anytime nature of these heuristic search algorithms, the search usually keeps going until an $\epsilon$-optimal action is found for the current belief $b_c$, or the available planning

```
 1: Function UPDATEANCESTORS(b')
    Inputs: b': An OR-Node for which we want to update all its ancestors.
    Static: b_c: The current belief state of the agent.
            T: An AND-OR tree representing the current search tree.
            L: A lower bound on V*.
            U: An upper bound on V*.
 2: while b' ≠ b_c do
 3:     Set (b, a) so that action a in belief b is parent node of belief node b'
 4:     L_T(b,a) ← R_B(b,a) + γ ∑_{z∈Z} Pr(z|b,a)L_T(τ(b,a,z))
 5:     U_T(b,a) ← R_B(b,a) + γ ∑_{z∈Z} Pr(z|b,a)U_T(τ(b,a,z))
 6:     z^T_{b,a} ← argmax_{z∈Z} H_T(b,a,z)H*_T(τ(b,a,z))
 7:     H*_T(b,a) ← H_T(b,a,z^T_{b,a})H*_T(τ(b,a,z^T_{b,a}))
 8:     b*_T(b,a) ← b*_T(τ(b,a,z^T_{b,a}))
 9:     L_T(b) ← max_{a'∈A} L_T(b,a')
10:     U_T(b) ← max_{a'∈A} U_T(b,a')
11:     a^T_b ← argmax_{a'∈A} H_T(b,a')H*_T(b,a')
12:     H*_T(b) ← H_T(b,a^T_b)H*_T(b,a^T_b)
13:     b*_T(b) ← b*_T(b,a^T_b)
14:     b' ← b
15: end while
```

**Algorithm 3.6:** UPDATEANCESTORS : Updates the bounds of the ancestors of all ancestors of an OR-Node

time is elapsed. An $\epsilon$-optimal action is found whenever $U_T(b_c) - L_T(b_c) \leq \epsilon$ or $L_T(b_c) \geq U_T(b_c, a'), \forall a' \neq \text{argmax}_{a \in A} L_T(b_c, a)$ (i.e. all other actions are pruned, in which case the optimal action has been found).

Now that we have covered the basic subroutines, we present the different heuristics proposed by Satia and Lave (1973), Washington (1997) and Ross and Chaib-draa (2007). We begin by introducing some useful notation.

Given any graph structure $G$, let us denote $\mathcal{F}(G)$ the set of all fringe nodes in $G$ and $\mathcal{H}_G(b, b')$ the set of all sequences of actions and observations that lead to belief node $b'$ from belief node $b$ in the search graph $G$. If we have a tree $T$, then $\mathcal{H}_T(b, b')$ will contain at most a single sequence which we will denote $h_T^{b,b'}$. Now given a sequence $h \in \mathcal{H}_G(b, b')$, we define $\Pr(h_z|b, h_a)$ the probability that we observe the whole sequence of observations $h_z$ in $h$, given that we start in belief node $b$ and perform the whole sequence of actions $h_a$ in $h$. Finally, we define $\Pr(h|b, \pi)$ to be the probability that we follow the entire action/observation sequence $h$ if we start in belief $b$ and behave according to policy $\pi$. Formally, these probabilities are computed as follows:

$$\Pr(h_z|b, h_a) = \prod_{i=1}^{d(h)} \Pr(h_z^i|b^{h_{i-1}}, h_a^i), \tag{25}$$

$$\Pr(h|b, \pi) = \prod_{i=1}^{d(h)} \Pr(h_z^i|b^{h_{i-1}}, h_a^i)\pi(b^{h_{i-1}}, h_a^i), \tag{26}$$

where $d(h)$ represents the depth of $h$ (number of actions in the sequence $h$), $h_a^i$ denotes the $i^{th}$ action in sequence $h$, $h_z^i$ the $i^{th}$ observation in the sequence $h$, and $b^{h_i}$ the belief state obtained by taking the first $i$ actions and observations of the sequence $h$ from $b$. Note that $b^{h_0} = b$.

### 3.4.1 Satia and Lave

The approach of Satia and Lave (1973) follows the heuristic search framework presented above. The main feature of this approach is to explore, at each iteration, the fringe node $b$ in the current search tree $T$ that maximizes the following term:

$$H_T(b_c, b) = \gamma^{d(h_T^{b_c, b})} \Pr(h_{T,z}^{b_c, b} | b_c, h_{T,a}^{b_c, b})(U_T(b) - L_T(b)),  \tag{27}$$

where $b \in \mathcal{F}(T)$ and $b_c$ is the root node of $T$. The intuition behind this heuristic is simple: recalling the definition of $V^*$, we note that the weight of the value $V^*(b)$ of a fringe node $b$ on $V^*(b_c)$ would be exactly $\gamma^{d(h_T^{b_c, b})} \Pr(h_{T,z}^{b_c, b} | b_c, h_{T,a}^{b_c, b})$, provided $h_{T,a}^{b_c, b}$ is a sequence of optimal actions. The fringe nodes where this weight is high have the most effect on the estimate of $V^*(b_c)$. Hence one should try to minimize the error at these nodes first. The term $U_T(b) - L_T(b)$ is included since it is an upper bound on the (unknown) error $V^*(b) - L_T(b)$. Thus this heuristic focuses the search in areas of the tree that most affect the value $V^*(b_c)$ and where the error is possibly large. This approach also uses Branch-and-Bound pruning, such that a fringe node reached by an action that is dominated in some parent belief $b$ is never going to be expanded. Using the same notation as in Algorithms 3.5 and 3.6, this heuristic can be implemented by defining $H_T(b)$, $H_T(b, a)$ and $H_T(b, a, z)$, as follows:

$$
\begin{aligned}
H_T(b) &= U_T(b) - L_T(b), \\
H_T(b, a) &= \begin{cases} 1 & \text{if } U_T(b, a) > L_T(b), \\ 0 & \text{otherwise,} \end{cases} \\
H_T(b, a, z) &= \gamma \Pr(z | b, a),
\end{aligned}
\tag{28}
$$

The condition $U_T(b, a) > L_T(b)$ ensures that the global heuristic value $H_T(b_c, b') = 0$ if some action in the sequence $h_{T,a}^{b_c, b'}$ is dominated (pruned). This guarantees that such fringe nodes will never be expanded.

Satia and Lave's heuristic focuses the search towards beliefs that are most likely to be reached in the future, and where the error is large. This heuristic is likely to be efficient in domains with a large number of observations, but only if the probability distribution over observations is concentrated over only a few observations. The term $U_T(b) - L_T(b)$ in the heuristic also prevents the search from doing unnecessary computations in areas of the tree where it already has a good estimate of the value function. This term is most efficient when the bounds computed offline, $U$ and $L$, are sufficiently informative. Similarly, node pruning is only going to be efficient if $U$ and $L$ are sufficiently tight, otherwise few actions will be pruned.

### 3.4.2 BI-POMDP

Washington (1997) proposed a slightly different approach inspired by the AO* algorithm (Nilsson, 1980), where the search is only conducted in the best solution graph. In the case of online POMDPs, this corresponds to the subtree of all belief nodes that are reached by sequences of actions maximizing the upper bound in their parent beliefs.

The set of fringe nodes in the best solution graph of $G$, which we denote $\widehat{\mathcal{F}}(G)$, can be defined formally as $\widehat{\mathcal{F}}(G) = \{b \in \mathcal{F}(G) | \exists h \in H_G(b_c, b), \Pr(h | b, \hat{\pi}_G) > 0\}$, where $\hat{\pi}_G(b, a) = 1$ if $a = \text{argmax}_{a' \in A} U_G(b, a')$ and $\hat{\pi}_G(b, a) = 0$ otherwise. The AO* algorithm simply specifies expanding any of these fringe nodes. Washington (1997) recommends exploring the fringe node in $\widehat{\mathcal{F}}(G)$ (where $G$ is the current acyclic search graph) that maximizes $U_G(b) - L_G(b)$. Washington's heuristic can be implemented by defining $H_T(b)$, $H_T(b,a)$ and $H_T(b,a,z)$, as follows:

$$
\begin{aligned}
H_T(b) &= U_T(b) - L_T(b), \\
H_T(b,a) &= \begin{cases} 1 & \text{if } a = \text{argmax}_{a' \in A} U_T(b,a'), \\ 0 & \text{otherwise}, \end{cases} \\
H_T(b,a,z) &= 1.
\end{aligned}
\tag{29}
$$

This heuristic tries to guide the search towards nodes that are reachable by "promising" actions, especially when they have loose bounds on their values (possibly large error). One nice property of this approach is that expanding fringe nodes in the best solution graph is the only way to reduce the upper bound at the root node $b_c$. This was not the case for Satia and Lave's heuristic. However, Washington's heuristic does not take into account the probability $\text{Pr}(h_z|b,h_a)$, nor the discount factor $\gamma^{d(h)}$, such that it may end up exploring nodes that have a very small probability of being reached in the future, and thus have little effect on the value of $V^*(b_c)$. Hence, it may not explore the most relevant nodes for optimizing the decision at $b_c$. This heuristic is appropriate when the upper bound $U$ computed offline is sufficiently informative, such that actions with highest upper bound would also usually tend to have highest Q-value. In such cases, the algorithm will focus its search on these actions and thus should find the optimal action more quickly then if it explored all actions equally. On the other hand, because it does not consider the observation probabilities, this approach may not scale well to large observation sets, as it will not be able to focus its search towards the most relevant observations.

### 3.4.3 AEMS

Ross and Chaib-draa (2007) introduced a heuristic that combines the advantages of BI-POMDP, and Satia and Lave's heuristic. It is based on a theoretical error analysis of tree search in POMDPs, presented by Ross et al. (2008).

The core idea is to expand the tree such as to reduce the error on $V^*(b_c)$ as quickly as possible. This is achieved by expanding the fringe node $b$ that contributes the most to the error on $V^*(b_c)$. The exact error contribution $e_T(b_c, b)$ of fringe node $b$ on $b_c$ in tree $T$ is defined by the following equation:

$$
e_T(b_c, b) = \gamma^{d(h_T^{b_c, b})} \text{Pr}(h_T^{b_c, b}|b_c, \pi^*)(V^*(b) - L_T(b)).
\tag{30}
$$

This expression requires $\pi^*$ and $V^*$ to be computed exactly. In practice, Ross and Chaib-draa (2007) suggest approximating the exact error $(V^*(b) - L_T(b))$ by $(U_T(b) - L_T(b))$, as was done by Satia and Lave, and Washington. They also suggest approximating $\pi^*$ by some policy $\hat{\pi}_T$, where $\hat{\pi}_T(b,a)$ represents the probability that action $a$ is optimal in its parent belief $b$, given its lower and upper bounds in tree $T$. In particular, Ross et al. (2008) considered two possible approximations for $\pi^*$. The first one is based on a uniformity assumption on the distribution of Q-values between the lower and upper bounds, which yields:

$$\hat{\pi}_T(b,a) = \begin{cases} \eta \frac{(U_T(b,a)-L_T(b))^2}{U_T(b,a)-L_T(b,a)} & \text{if } U_T(b,a) > L_T(b), \\ 0 & \text{otherwise,} \end{cases} \tag{31}$$

where $\eta$ is a normalization constant such that the sum of the probabilities $\hat{\pi}_T(b,a)$ over all actions equals 1.

The second is inspired by AO* and BI-POMDP, and assumes that the action maximizing the upper bound is in fact the optimal action:

$$\hat{\pi}_T(b,a) = \begin{cases} 1 & \text{if } a = \operatorname{argmax}_{a' \in A} U_T(b,a'), \\ 0 & \text{otherwise.} \end{cases} \tag{32}$$

Given the approximation $\hat{\pi}_T$ of $\pi^*$, the AEMS heuristic will explore the fringe node $b$ that maximizes:

$$H_T(b_c, b) = \gamma^{d(h_T^{b_c, b})} \Pr(h_T^{b_c, b} | b_c, \hat{\pi}_T)(U_T(b) - L_T(b)). \tag{33}$$

This can be implemented by defining $H_T(b)$, $H_T(b,a)$ and $H_T(b,a,z)$ as follows:

$$\begin{aligned} H_T(b) &= U_T(b) - L_T(b), \\ H_T(b,a) &= \hat{\pi}_T(b,a), \\ H_T(b,a,z) &= \gamma \Pr(z|b,a). \end{aligned} \tag{34}$$

We refer to this heuristic as AEMS1 when $\hat{\pi}_T$ is defined as in Equation 31, and AEMS2 when it is defined as in Equation 32.[2]

Let us now examine how AEMS combines the advantages of both the Satia and Lave, and BI-POMDP heuristics. First, AEMS encourages exploration of nodes with loose bounds and possibly large error by considering the term $U_T(b) - L_T(b)$ as in previous heuristics. Moreover, as in Satia and Lave, it focuses the exploration towards belief states that are likely to be encountered in the future. This is good for two reasons. As mentioned before, if a belief state has a low probability of occurrence in the future, it has a limited effect on the value $V^*(b_c)$ and thus it is not necessary to know its value precisely. Second, exploring the highly probable belief states increases the chance that we will be able to reuse these computations in the future. Hence, AEMS should be able to deal efficiently with large observation sets, assuming the distribution over observations is concentrated over a few observations. Finally, as in BI-POMDP, AEMS favors the exploration of fringe nodes reachable by actions that seem more likely to be optimal (according to $\hat{\pi}_T$). This is useful to handle large action sets, as it focuses the search on actions that look promising. If these "promising" actions are not optimal, then this will quickly become apparent. This will work well if the best actions have the highest probabilities in $\hat{\pi}_T$. Furthermore, it is possible to define $\hat{\pi}_T$ such that it automatically prunes dominated actions by ensuring that $\hat{\pi}_T(b,a) = 0$ whenever $U_T(b,a) < L_T(b)$. In such cases, the heuristic will never choose to expand a fringe node reached by a dominated action.

As a final note, Ross et al. (2008) determined sufficient conditions under which the search algorithm using this heuristic is guaranteed to find an $\epsilon$-optimal action within finite time. This is stated in Theorem 3.2.

---

2. The AEMS2 heuristic was also used for a policy search algorithm by Hansen (1998).

**Theorem 3.2.** *(Ross et al., 2008) Let $\epsilon > 0$ and $b_c$ the current belief. If for any tree $T$ and parent belief $b$ in $T$ where $U_T(b) - L_T(b) > \epsilon$, $\hat{\pi}_T(b,a) > 0$ for $a = \text{argmax}_{a' \in A} U_T(b,a')$, then the AEMS algorithm is guaranteed to find an $\epsilon$-optimal action for $b_c$ within finite time.*

We observe from this theorem that it is possible to define many different policies $\hat{\pi}_T$ under which the AEMS heuristic is guaranteed to converge. AEMS1 and AEMS2 both satisfy this condition.

### 3.4.4 HSVI

A heuristic similar to AEMS2 was also used by Smith and Simmons (2004) for their offline value iteration algorithm HSVI as a way to pick the next belief point at which to perform $\alpha$-vector backups. The main difference is that HSVI proceeds via a greedy search that descends the tree from the root node $b_0$, going down towards the action that maximizes the upper bound and then the observation that maximizes $\Pr(z|b,a)(U(\tau(b,a,z)) - L(\tau(b,a,z)))$ at each level, until it reaches a belief $b$ at depth $d$ where $\gamma^{-d}(U(b) - L(b)) < \epsilon$. This heuristic could be used in an online heuristic search algorithm by instead stopping the greedy search process when it reaches a fringe node of the tree and then selecting this node as the one to be expanded next. In such a setting, HSVI's heuristic would return a greedy approximation of the AEMS2 heuristic, as it may not find the fringe node which actually maximizes $\gamma^{d(h_T^{b_c,b})} \Pr(h_T^{b_c,b}|b_c, \hat{\pi}_T)(U_T(b) - L_T(b))$. We consider this online version of the HSVI heuristic in our empirical study (Section 4). We refer to this extension as HSVI-BFS. Note that the complexity of this greedy search is the same as finding the best fringe node via the dynamic programming process that updates $H_T^*$ and $b_T^*$ in the UPDATEANCESTORS subroutine.

## 3.5 Alternatives to Tree Search

We now present two alternative online approaches that do not proceed via a lookahead search in the belief MDP. In all online approaches presented so far, one problem is that no learning is achieved over time, i.e. everytime the agent encounters the same belief, it has to recompute its policy starting from the same initial upper and lower bounds computed offline. The two online approaches presented next address this problem by presenting alternative ways of updating the initial value functions computed offline so that the performance of the agent improves over time as it stores updated values computed at each time step. However, as is argued below and in the discussion (Section 5.2), these techniques lead to other disadvantages in terms of memory consumption and/or time complexity.

### 3.5.1 RTDP-BEL

An alternative approach to searching in AND-OR graphs is the RTDP algorithm (Barto et al., 1995) which has been adapted to solve POMDPs by Geffner and Bonet (1998). Their algorithm, called RTDP-BEL, learns approximate values for the belief states visited by successive trials in the environment. At each belief state visited, the agent evaluates all possible actions by estimating the expected reward of taking action $a$ in the current belief

```
1: Function ONLINEPOMDPSOLVER()
   Static: b_c: The current belief state of the agent.
           V_0: Initial approximate value function (computed offline).
           V: A hashtable of beliefs and their approximate value.
           k: Discretization resolution.

2: Initialize b_c to the initial belief state and V to an empty hashtable.
3: while not EXECUTIONTERMINATED() do
4:    For all a ∈ A: Evaluate Q(b_c, a) = R_B(b_c, a) + γ ∑_{z∈Z} Pr(z|b, a)V(DISCRETIZE(τ(b, a, z), k))
5:    â ← argmax_{a∈A} Q(b_c, a)
6:    Execute best action â for b_c
7:    V(DISCRETIZE(b_c, k)) ← Q(b_c, â)
8:    Perceive a new observation z
9:    b_c ← τ(b_c, â, z)
10: end while
```

**Algorithm 3.7:** RTDP-Bel Algorithm.

state $b$ with an approximate Q-value equation:

$$Q(b, a) = R_B(b, a) + γ \sum_{z∈Z} Pr(z|b, a)V(τ(b, a, z)), \qquad (35)$$

where $V(b)$ is the value learned for the belief $b$.

If the belief state $b$ has no value in the table, then it is initialized to some heuristic value. The authors suggest using the MDP approximation for the initial value of each belief state. The agent then executes the action that returned the greatest $Q(b, a)$ value. Afterwards, the value $V(b)$ in the table is updated with the $Q(b, a)$ value of the best action. Finally, the agent executes the chosen action and it makes the new observation, ending up in a new belief state. This process is then repeated again in this new belief.

The RTDP-BEL algorithm learns a heuristic value for each belief state visited. To maintain an estimated value for each belief state in memory, it needs to discretize the belief state space to have a finite number of belief states. This also allows generalization of the value function to unseen belief states. However, it might be difficult to find the best discretization for a given problem. In practice, this algorithm needs substantial amounts of memory (greater than 1GB in some cases) to store all the learned belief state values, especially in POMDPs with large state spaces. The implementation of the RTDP-Bel algorithm is presented in Algorithm 3.7.

The function DISCRETIZE($b, k$) returns a discretized belief $b'$ where $b'(s) = \text{round}(kb(s))/k$ for all states $s ∈ S$, and $V(b)$ looks up the value of belief $b$ in a hashtable. If $b$ is not present in the hashtable, the value $V_0(b)$ is returned by $V$. Supported by experimental data, Geffner and Bonet (1998) suggest choosing $k ∈ [10, 100]$, as it usually produces the best results. Notice that for a discretization resolution of $k$ there are $O((k+1)^{|S|})$ possible discretized beliefs. This implies that the memory storage required to maintain $V$ is exponential in $|S|$, which becomes quickly intractable, even for mid-size problems. Furthermore, learning good estimates for this exponentially large number of beliefs usually requires a very large number of trials, which might be infeasible in practice. This technique can sometimes be applied in large domains when a factorized representation is available. In such cases, the belief can be maintained as a set of distributions (one for subset of conditionaly independent state variables) and the discretization applied seperately to each distribution. This can greatly reduce the possible number of discretized beliefs.

| Algorithm | $\epsilon$-optimal | Anytime | Branch & Bound | Monte Carlo | Heuristic | Learning |
|---|---|---|---|---|---|---|
| RTBSS | yes | no | yes | no | no | no |
| McAllester | high probability | no | no | yes | no | no |
| Rollout | no | no | no | yes | no | no |
| Satia and Lave | yes | yes | yes | no | yes | no |
| Washington | acyclic graph | yes | implicit | no | yes | no |
| AEMS | yes | yes | implicit | no | yes | no |
| HSVI-BFS | yes | yes | implicit | no | yes | no |
| RTDP-Bel | no | no | no | no | no | yes |
| SOVI | yes | yes | no | no | no | yes |

Table 1: Properties of various online methods.

### 3.5.2 SOVI

A more recent online approach, called SOVI (Shani et al., 2005), extends HSVI (Smith & Simmons, 2004, 2005) into an online value iteration algorithm. This approach maintains a priority queue of the belief states encountered during the execution and proceeds by doing $\alpha$-vector updates for the current belief state and the $k$ belief states with highest priority at each time step. The priority of a belief state is computed according to how much the value function changed at successor belief states, since the last time it was updated. Its authors also propose other improvements to the HSVI algorithm to improve scalability, such as a more efficient $\alpha$-vector pruning technique, and avoiding to use linear programs to update and evaluate the upper bound. The main drawback of this approach is that it is hardly applicable in large environments with short real-time constraints, since it needs to perform a value iteration update with $\alpha$-vectors online, and this can have very high complexity as the number of $\alpha$-vectors representing the value function increases (i.e. $O(k|S||A||Z|(|S| + |\Gamma_{t-1}|))$ to compute $\Gamma_t$).

### 3.6 Summary of Online POMDP Algorithms

In summary, we see that most online POMDP approaches are based on lookahead search. To improve scalability, different techniques are used: branch-and-bound pruning, search heuristics, and Monte Carlo sampling. These techniques reduce the complexity from different angles. Branch-and-bound pruning lowers the complexity related to the action space size. Monte Carlo sampling has been used to lower the complexity related to the observation space size, and could also potentially be used to reduce the complexity related to the action space size (by sampling a subset of actions). Search heuristics lower the complexity related to actions and observations by orienting the search towards the most relevant actions and observations. When appropriate, factored POMDP representations can be used to reduce the complexity related to state. A summary of the different properties of each online algorithm is presented in Table 1.

## 4. Empirical Study

In this section, we compare several online approaches in two domains found in the POMDP literature: *Tag* (Pineau et al., 2003) and *RockSample* (Smith & Simmons, 2004). We consider a modified version of *RockSample*, called *FieldVisionRockSample* (Ross & Chaib-draa, 2007), that has a higher observation space than the original *RockSample*. This environment is introduced as a means to test and compare the different algorithms in environments with large observation spaces.

### 4.1 Methodology

For each environment, we first compare the real-time performance of the different heuristics presented in Section 3.4 by limiting their planning time to 1 second per action. All heuristics were given the same lower and upper bounds such that their results would be comparable. The objective here is to evaluate which search heuristic is most efficient in different types of environments. To this end, we have implemented the different search heuristics (Satia and Lave, BI-POMDP, HSVI-BFS and AEMS) into the same best-first search algorithm, such that we can directly measure the efficiency of the heuristic itself. Results were also obtained for different lower bounds (Blind and PBVI) to verify how this choice affects the heuristic's efficiency. Finally, we compare how online and offline times affect the performance of each approach. Except where stated otherwise, all experiments were run on an Intel Xeon 2.4 Ghz with 4GB of RAM; processes were limited to 1GB of RAM.

#### 4.1.1 Metrics to compare online approaches

We compare performance first and foremost in terms of average discounted return at execution time. However, what we really seek with online approaches is to guarantee better solution quality than that provided by the original bounds. In other words, we seek to reduce the error of the original bounds as much as possible. This suggests that a good metric for the efficiency of online algorithms is to compare the improvement in terms of the error bounds at the current belief before and after the online search. Hence, we define the error bound reduction percentage to be:

$$\text{EBR}(b) = 1 - \frac{U_T(b) - L_T(b)}{U(b) - L(b)}, \tag{36}$$

where $U_T(b)$, $L_T(b)$, $U(b)$ and $L(b)$ are defined as in Section 3.2. The best online algorithm should provide the highest error bound reduction percentage, given the same initial bounds and real-time constraint.

Because the EBR metric does not necessarily reflect true error reduction, we also compare the return guarantees provided by each algorithm, i.e. the lower bounds on the expected return provided by the computed policies for the current belief. Because improvement of the lower bound compared to the initial lower bound computed offline is a direct indicator of true error reduction, the best online algorithm should provide the greatest lower bound improvement at the current belief, given the same initial bounds and real-time constraint. Formally, we define the lower bound improvement to be:

$$\text{LBI}(b) = L_T(b) - L(b). \tag{37}$$

In our experiments, both the EBR and LBI metrics are evaluated at each time step for the current belief. We are interested in seeing which approach provides the highest EBR and LBI on average.

We also consider other metrics pertaining to complexity and efficiency. In particular, we report the average number of belief nodes maintained in the search tree. Methods that have lower complexity will generally be able to maintain bigger trees, but the results will show that this does not always relate to higher error bound reduction and returns. We will also measure the efficiency of reusing part of the search tree by recording the percentage of belief nodes that were reused from one time step to the next.

### 4.2 *Tag*

*Tag* was initially introduced by Pineau et al. (2003). This environment has also been used more recently in the work of several authors (Poupart & Boutilier, 2003; Vlassis & Spaan, 2004; Pineau, 2004; Spaan & Vlassis, 2004; Smith & Simmons, 2004; Braziunas & Boutilier, 2004; Spaan & Vlassis, 2005; Smith & Simmons, 2005). For this environment, an approximate POMDP algorithm is necessary because of its large size (870 states, 5 actions and 30 observations). The *Tag* environment consists of an agent that has to catch (Tag) another agent while moving in a 29-cell grid domain. The reader is referred to the work of Pineau et al. (2003) for a full description of the domain. Note that for all results presented below, the belief state is represented in factored form. The domain is such that an exact factorization is possible.

To obtain results in *Tag*, we run each algorithm in each starting configuration 5 times, ( i.e. 5 runs for each of the 841 different starting joint positions, excluding the 29 terminal states ). The initial belief state is the same for all runs and consists of a uniform distribution over the possible joint agent positions.

Table 2 compares the different heuristics by presenting 95% confidence intervals on the average discounted return per run (Return), average error bound reduction percentage per time step (EBR), average lower bound improvement per time step (LBI), average belief nodes in the search tree per time step (Belief Nodes), the average percentage of belief nodes reused per time step (Nodes Reused), the average online planning time used per time step (Online Time). In all cases, we use the FIB upper bound and the Blind lower bound. Note that the average online time is slightly lower than 1 second per step because algorithms sometimes find $\epsilon$-optimal solutions in less than a second.

We observe that the efficiency of HSVI-BFS, BI-POMDP and AEMS2 differs slightly in this environment and that they outperform the three other heuristics: RTBSS, Satia and Lave, and AEMS1. The difference can be explained by the fact that the latter three methods do not restrict the search to the best solution graph. As a consequence, they explore many irrelevant nodes, as shown by the lower error bound reduction percentage, lower bound improvement, and nodes reused. This poor reuse percentage explains why Satia and Lave, and AEMS1 were limited to a lower number of belief nodes in their search tree, compared to the other methods which reached averages around 70K. The results of the three other heuristics do not differ much here because the three heuristics only differ in the way they choose the observations to explore in the search. Since only two observations are possible after the first action and observation, and one of these observations leads directly to a terminal belief state, the possibility that the heuristics differed significantly was very limited. Due to this limitation of the Tag domain, we now compare these online algorithms in a larger and more complex domain: RockSample.

| Heuristic | Return | EBR (%) | LBI | Belief Nodes | Nodes Reused (%) | Online Time (ms) |
|---|---|---|---|---|---|---|
| RTBSS(5) | -10.31 ± 0.22 | 22.3 ± 0.4 | 3.03 ± 0.07 | 45066 ± 701 | 0 | 580 ± 9 |
| Satia and Lave | -8.35 ± 0.18 | 22.9 ± 0.2 | 2.47 ± 0.04 | 36908 ± 209 | 10.0 ± 0.2 | 856 ± 4 |
| AEMS1 | -6.73 ± 0.15 | 49.0 ± 0.3 | 3.92 ± 0.03 | 43693 ± 314 | 25.1 ± 0.3 | 814± 4 |
| HSVI-BFS | -6.22 ± 0.19 | 75.7 ± 0.4 | 7.69 ± 0.06 | 64870 ± 947 | 54.1 ± 0.7 | 673 ± 5 |
| BI-POMDP | -6.22 ± 0.15 | 76.2 ± 0.5 | 7.81 ± 0.06 | 79508 ± 1000 | 54.6 ± 0.6 | 622 ± 4 |
| AEMS2 | -6.19 ± 0.15 | 76.3 ± 0.5 | 7.81 ± 0.06 | 80250 ± 1018 | 54.8 ± 0.6 | 623 ± 4 |

Table 2: Comparison of different search heuristics on the *Tag* environment using the Blind policy as a lower bound.

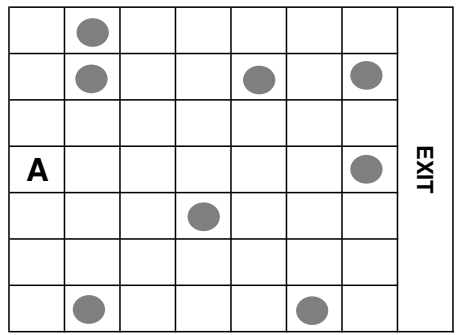

Figure 3: *RockSample[7,8]*.

### 4.3 *RockSample*

The *RockSample* problem was originally presented by Smith and Simmons (2004). In this domain, an agent has to explore the environment and sample some rocks (see Figure 3), similarly to what a real robot would do on the planet Mars. The agent receives rewards by sampling rocks and by leaving the environment (at the extreme right of the environment). A rock can have a scientific value or not, and the agent has to sample only good rocks.

We define *RockSample[n, k]* as an instance of the *RockSample* problem with an $n \times n$ grid and $k$ rocks. A state is characterized by $k+1$ variables: $X_P$, which defines the position of the robot and can take values $\{(1,1),(1,2),\ldots,(n,n)\}$ and $k$ variables, $X_1^R$ through $X_k^R$, representing each rock, which can take values $\{Good, Bad\}$.

The agent can perform $k+5$ actions: $\{North, South, East, West, Sample, Check_1, \ldots, Check_k\}$. The four motion actions are deterministic. The *Sample* action samples the rock at the agent's current location. Each $Check_i$ action returns a noisy observation from $\{Good, Bad\}$ for rock $i$.

The belief state is represented in factored form by the known position and a set of $k$ probabilities, namely the probability of each rock being good. Since the observation of a rock state is independent of the other rock states (it only depends on the known robot position), the complexity of computing $\Pr(z|b,a)$ and $\tau(b,a,z)$ is greatly reduced. Effectively, the computation of $\Pr(z|b, Check_i)$ reduces to: $\Pr(z|b, Check_i) = \Pr(Accurate|X_P, Check_i) \cdot \Pr(X_i^R = z) + (1 - \Pr(Accurate|X_P, Check_i)) \cdot (1 - \Pr(X_i^R = z))$. The probability that the sensor is accurate on rock $i$, $\Pr(Accurate|X_P, Check_i) = \frac{1+\eta(X_p,i)}{2}$, where $\eta(X_p, i) = 2^{-d(X_p,i)/d_0}$, $d(X_p, i)$ is the euclidean distance between position $X_p$ and the position of rock $i$, and $d_0$ is a constant specifying the *half efficiency distance*. $\Pr(X_i^R = z)$ is obtained directly from the probability (stored in $b$) that rock $i$ is good. Similarly, $\tau(b, a, z)$ can be computed quite easily as the move actions deterministically affect variable $X_P$, and a $Check_i$ action only changes the probability associated to $X_i^R$ according to the sensor's accuracy.

| Heuristic | Return | EBR (%) | LBI | Belief Nodes | Nodes Reused (%) | Online Time (ms) |
|---|---|---|---|---|---|---|
| Blind: Return:7.35, $|\Gamma| = 1$, Time:4s | | | | | | |
| Satia and Lave | $7.35 \pm 0$ | $3.64 \pm 0$ | $0 \pm 0$ | $509 \pm 0$ | $8.92 \pm 0$ | $900 \pm 0$ |
| AEMS1 | $10.30 \pm 0.08$ | $9.50 \pm 0.11$ | $0.90 \pm 0.03$ | $579 \pm 2$ | $5.31 \pm 0.03$ | $916 \pm 1$ |
| RTBSS(2) | $10.30 \pm 0.15$ | $9.65 \pm 0.02$ | $1.00 \pm 0.04$ | $439 \pm 0$ | $0 \pm 0$ | $886 \pm 2$ |
| BI-POMDP | $18.43 \pm 0.14$ | $33.3 \pm 0.5$ | $4.33 \pm 0.06$ | $2152 \pm 71$ | $29.9 \pm 0.6$ | $953 \pm 2$ |
| HSVI-BFS | $20.53 \pm 0.31$ | $51.7 \pm 0.7$ | $5.25 \pm 0.07$ | $2582 \pm 72$ | $36.5 \pm 0.5$ | $885 \pm 5$ |
| AEMS2 | $20.75 \pm 0.15$ | $52.4 \pm 0.6$ | $5.30 \pm 0.06$ | $3145 \pm 101$ | $36.4 \pm 0.5$ | $859 \pm 6$ |
| PBVI: Return:5.93, $|B| = 64$, $|\Gamma| = 54$, Time:2418s | | | | | | |
| AEMS1 | $17.10 \pm 0.28$ | $26.1 \pm 0.4$ | $1.39 \pm 0.03$ | $1461 \pm 28$ | $12.2 \pm 0.1$ | $954 \pm 2$ |
| Satia and Lave | $19.09 \pm 0.21$ | $16.9 \pm 0.1$ | $1.17 \pm 0.01$ | $2311 \pm 25$ | $13.5 \pm 0.1$ | $965 \pm 1$ |
| RTBSS(2) | $19.45 \pm 0.30$ | $22.4 \pm 0.3$ | $1.37 \pm 0.04$ | $426 \pm 1$ | $0 \pm 0$ | $540 \pm 7$ |
| BI-POMDP | $21.36 \pm 0.22$ | $49.5 \pm 0.2$ | $2.73 \pm 0.02$ | $2781 \pm 38$ | $32.2 \pm 0.2$ | $892 \pm 2$ |
| AEMS2 | $21.37 \pm 0.22$ | $57.7 \pm 0.2$ | $3.08 \pm 0.02$ | $2910 \pm 46$ | $38.2 \pm 0.2$ | $826 \pm 3$ |
| HSVI-BFS | $21.46 \pm 0.22$ | $56.3 \pm 0.2$ | $3.03 \pm 0.02$ | $2184 \pm 33$ | $37.3 \pm 0.2$ | $826 \pm 2$ |

Table 3: Comparison of different search heuristics in *RockSample[7,8]* environment, using the Blind policy and PBVI as a lower bound.

To obtain our results in RockSample, we run each algorithm in each starting rock configuration 20 times (i.e. 20 runs for each of the $2^k$ different joint rock states). The initial belief state is the same for all these runs and consists of 0.5 that each rock is good, plus the known initial robot position.

### 4.3.1 Real-Time Performance of Online Search

In Table 3, we present 95% confidence intervals on the mean of our metrics of interest, for RockSample[7,8] (12545 states, 13 actions, 2 observations), with real-time contraints of 1 second per action. We compare performance using two different lower bounds, the Blind policy and PBVI, and use QMDP for the upper bound in both cases. The performance of the policy defined by each lower bound is shown in the comparison header. For RTBSS, the notation RTBSS($k$) indicates a $k$-step lookahead; we use the depth $k$ that yields an average online time closest to 1 second per action.

**Return** In terms of the return, we first observe that the AEMS2 and HSVI-BFS heuristics obtain very similar results. Each of these obtains the highest return by a slight margin with one of the lower bounds. BI-POMDP obtains a similar return when combined with the PBVI lower bound, but performs much worse with the Blind lower bound. The two other heuristics, Satia and Lave, and AEMS1, perform considerably worse in terms of return with either lower bound.

**EBR and LBI**  In terms of error bound reduction and lower bound improvement, AEMS2 obtains the best results with both lower bounds. HSVI-BFS is a close second. This indicates that AEMS2 can more effectively reduce the true error than the other heuristics, and therefore, guarantees better performance. While BI-POMDP tends to be less efficient than AEMS2 and HSVI-BFS, it does significantly better than RTBSS, Satia and Lave, and AEMS1, which only slightly improve the bounds in both case. Satia and Lave is unable to increase the Blind lower bound, which explains why it obtains the same return as the Blind policy. We also observe that the higher the error bound reduction and lower bound improvement, the higher the average discounted return usually is. This confirms our intuition that guiding the search such as to minimize the error at the current belief $b_c$ is a good strategy to obtain better return.

**Nodes Reused**  In terms of the percentage of nodes reused, AEMS2 and HVSI-BFS generally obtain the best scores. This allows these algorithms to maintain a higher number of nodes in their trees, which could also partly explain why they outperform the other heuristics in terms of return, error bound reduction and lower bound improvement. Note that RTBSS does not reuse any node in the tree because the algorithm does not store the tree in memory. As a consequence, the reuse percentage is always 0.

**Online Time**  Finally, we also observe that AEMS2 requires less average online time per action than the other algorithms to attain its performance. In general, a lower average online time means the heuristic is efficient at finding $\epsilon$-optimal actions in a small amount of time. The running time for RTBSS is determined by the chosen depth, as it cannot stop before completing the full lookahead search.

**Summary**  Overall, we see that AEMS2 and HSVI-BFS obtain similar results. However AEMS2 seems slightly better than HSVI-BFS, as it provides better performance guarantees (lower error) within a shorter period of time. But the difference is not very significant. This may be due to the small number of observations in this environment, in which case the two heuristics expand the tree in very similar ways. In the next section, we explore a domain with many more observations to evaluate the impact of this factor.

The lower performances of the three other heuristics can be explained by various reasons. In the case of BI-POMDP, this is due to the fact that it does not take into account the observation probabilities $\Pr(z|b, a)$ and discount factor $\gamma$ in the heuristic value. Hence it tend to expand fringe nodes that did not affect significantly the value of the current belief. As for Satia and Lave, its poor performance in the case of the Blind policy can be explained by the fact that the fringe nodes that maximize this heuristic are always leaves reached by a sequence of *Move* actions. Due to the deterministic nature of the *Move* actions ($\Pr(z|b, a) = 1$ for these actions, whereas *Check* actions have $\Pr(z|b, a) = 0.5$ initially), the heuristic value for fringe nodes reached by *Move* actions is much higher until the error is reduced significantly. As a result, the algorithm never explores any nodes under the *Check* actions, and the robot always follows the Blind policy (moving east, never checking or sampling any rocks). This demonstrates the importance of restricting the choice of which leaves to explore to those reached by a sequence of actions maximizing the upper bound, as done in AEMS2, HSVI-BFS and BI-POMDP. In the case of AEMS1, it probably behaves less efficiently because the term it uses to estimate the probability that a certain action is optimal is not a good approximation in this environment. Moreover, because AEMS1 does not restrict the exploration to the best solution graph, it probably also suffers, in part, from the same problems as the Satia and Lave heuristic. RTBSS also did not perform very well with the Blind lower bound. This is due to the short depth allowed to search the tree, required to have a running time $\leq 1$ second/action. This confirms that we can do significantly better than an exhaustive search by having good heuristics to guide the search.

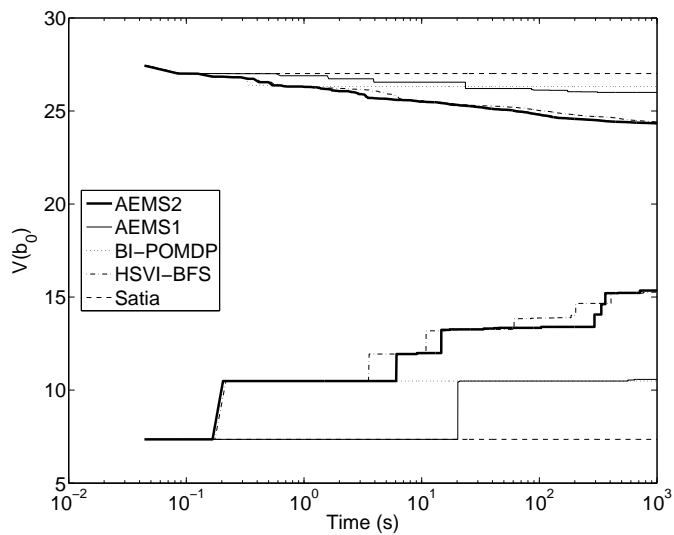

Figure 4: Evolution of the upper and lower bounds on *RockSample[7,8]*.

### 4.3.2 Long-Term Error Reduction of Online Heuristic Search

To compare the long term performance of the different heuristics, we let the algorithms run in offline mode from the initial belief state of the environment, and log changes in the lower and upper bound values of this initial belief state over 1000 seconds. Here, the initial lower and upper bounds are provided by the Blind policy and QMDP respectively. We see from Figure 4 that Satia and Lave, AEMS1 and BI-POMDP are not as efficient as HSVI-BFS and AEMS2 at reducing the error on the bounds. One interesting thing to note is that the upper bound tends to decrease slowly but continuously, whereas the lower bound often increases in a stepwise manner. We believe this is due to the fact that the upper bound is much tighter than the lower bound. We also observe that most of the error bound reduction happens in the first few seconds of the search. This confirms that the nodes expanded earlier in the tree have much more impact on the error of $b_c$ than those expanded very far in the tree (e.g. after hundreds of seconds). This is an important result in support of using online (as opposed to offline) methods.

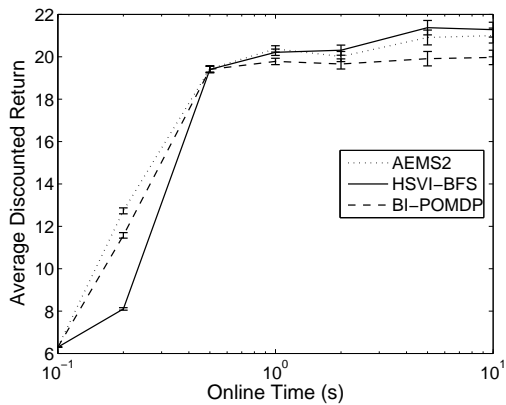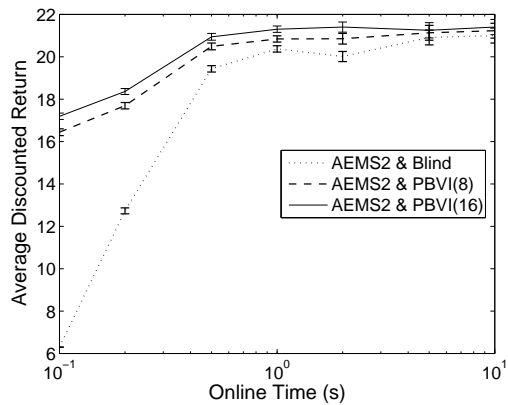

Figure 5: Comparison of the return as a function of the online time in RockSample(10,10) for different online methods.

Figure 6: Comparison of the return as a function of the online time in RockSample(10,10) for different offline lower bounds.

### 4.3.3 INFLUENCE OF OFFLINE AND ONLINE TIME

We now compare how the performance of online approaches is influenced by the available online and offline times. This allows us to verify if a particular method is better when the available online time is shorter (or longer), or whether increasing the offline time could be beneficial.

We consider the three approaches that have shown best overall performance so far (BI-POMDP, HSVI-BFS and AEMS2) and compare their average discounted return as a function of the online time constraint per action. Experiments were run in RockSample[10,10] (102,401 states, 15 actions, 2 observations) for each of the following online time constraints: 0.1s, 0.2s, 0.5s, 1s, 2s, 5s and 10s. To vary the offline time, we used 3 different lower bounds: Blind policy, PBVI with 8 belief points, and PBVI with 16 belief points, taking respectively 15s, 82s, and 193s. The upper bound used is QMDP in all cases. These results were obtained on an Intel Xeon 3.0 Ghz processor.

In Figure 5, we observe that AEMS2 fares significantly better than HSVI-BFS and BI-POMDP for short time constraints. As the time constraint increases, AEMS2 and HSVI-BFS performs similarly (no significant statistical difference). We also notice that the performance of BI-POMDP stops improving after 1 second of planning time. This can be explained by the fact that it does not take into account the observation probabilities $\Pr(z|b, a)$, nor the discount factor. As the search tree grows bigger, more and more fringe nodes have small probability of being reached in the future, such that it becomes more and more important to take these probabilities into account in order to improve performance. Otherwise, as we observe in the case of BI-POMDP, most expanded nodes do not affect the quality of the solution found.

From Figure 6, we observe that increasing the offline time has a beneficial effect mostly when we have very short real-time constraints. When more online planning time is available, the difference between the performances of AEMS2 with the Blind lower bound, and AEMS2 with PBVI becomes insignificant. However, for online time constraints smaller than one second, the difference in performance is very large. Intuitively, with very short real-time constraints the algorithm does not have enough time to expand a lot of nodes, such that the policy found relies much more on the bounds computed offline. On the other hand, with longer time constraints, the algorithm has enough time to significantly improve the bounds computed offline, and thus the policy found does not rely as much on the offline bounds.

### 4.4 *FieldVisionRockSample*

It seems from the results presented thus far that HSVI-BFS and AEMS2 have comparable performance on standard domains. We note however that these environments have very small observation sets (assuming observations with zero probability are removed). We believe AEMS2 is especially well suited for domains with large observation spaces. However, there are few such standard problems in the literature. We therefore consider a modified version of the *RockSample* environment, called *FieldVisionRockSample* (Ross & Chaib-draa, 2007), which has an observation space size exponential in the number of rocks.

The *FieldVisionRockSample* (FVRS) problem differs from the *RockSample* problem only in the way the robot is able to perceive the rocks in the environment. Recall that in *RockSample*, the agent has to do a *Check* action on a specific rock to observe its state through a noisy sensor. In FVRS, the robot observes the state of all rocks, through the same noisy sensor, after any action is conducted in the environment. Consequently, this eliminates the use of *Check* actions, and the remaining actions for the robot include only the four move actions {*North, East, South, West*} and the *Sample* action. The robot can perceive each rock as being either *Good* or *Bad*, thus the observation space size is $2^k$ for an instance of the problem with $k$ rocks. As in *RockSample*, the efficiency of the sensor is defined through the parameter $\eta = 2^{-d/d_0}$, where $d$ is the distance of the rock and $d_0$ is the *half efficiency distance*. We assume the sensor's observations are independent for each rock.

In FVRS, the partial observability of the environment is directly proportional to the parameter $d_0$: as $d_0$ increases, the sensor becomes more accurate and the uncertainty on the state of the environment decreases. The value $d_0$ defined for the different instances of *RockSample* in the work of Smith and Simmons (2004) is too high for the FVRS problem (especially in the bigger instances of *RockSample*), making it almost completely observable. Consequently, we re-define the value $d_0$ for the different instances of the *FieldVisionRock-Sample* according to the size of the grid $(n)$. By considering the fact that in an $n \times n$ grid, the largest possible distance between a rock and the robot is $(n-1)\sqrt(2)$, it seems reasonable that at this distance, the probability of observing the real state of the rock should be close to 50% for the problem to remain partially observable. Consequently, we define $d_0 = (n-1)\sqrt(2)/4$.

To obtain results for the FVRS domain, we run each algorithm in each starting rock configurations 20 times (i.e. 20 runs for each of the $2^k$ different joint rock states). The initial belief state is the same for all runs and corresponds to a probability of 0.5 that each rock is good, as well as the known initial position of the robot.

| Heuristic | Return | EBR (%) | LBI | Belief Nodes | Nodes Reused (%) | Online Time (ms) |
|---|---|---|---|---|---|---|
| FVRS[5,5]  [Blind: Return:8.15, \|Γ\| = 1, Time=170ms] | | | | | | |
| RTBSS(2) | 16.54 ± 0.37 | 18.4 ± 1.1 | 2.80 ± 0.19 | 18499 ± 102 | 0 ± 0 | 3135 ± 27 |
| AEMS1 | 16.88 ± 0.36 | 17.1 ± 1.1 | 2.35 ± 0.16 | 8053 ± 123 | 1.19 ± 0.07 | 876 ± 5 |
| Satia and Lave | 18.68 ± 0.39 | 15.9 ± 1.2 | 2.17 ± 0.16 | 7965 ± 118 | 0.88 ± 0.06 | 878 ± 4 |
| HSVI-BFS | 20.27 ± 0.44 | 23.8 ± 1.4 | 2.64 ± 0.14 | 4494 ± 105 | 4.50 ± 0.80 | 857 ± 12 |
| AEMS2 | 21.18 ± 0.45 | 31.5 ± 1.5 | 3.11 ± 0.15 | 12301 ± 440 | 3.93 ± 0.22 | 854 ± 13 |
| BI-POMDP | 22.75 ± 0.47 | 31.1 ± 1.2 | 3.30 ± 0.17 | 12199 ± 427 | 2.26 ± 0.44 | 782 ± 12 |
| FVRS[5,7]  [Blind: Return:8.15, \|Γ\| = 1, Time=761ms] | | | | | | |
| RTBSS(1) | 20.57 ± 0.23 | 7.72 ± 0.13 | 2.07 ± 0.11 | 516 ± 1 | 0 ± 0 | 254 ± 1 |
| BI-POMDP | 22.75 ± 0.25 | 11.1 ± 0.4 | 2.08 ± 0.07 | 4457 ± 61 | 0.37 ± 0.11 | 923 ± 2 |
| Satia and Lave | 22.79 ± 0.25 | 11.1 ± 0.4 | 2.05 ± 0.08 | 3683 ± 52 | 0.36 ± 0.07 | 947 ± 3 |
| AEMS1 | 23.31 ± 0.25 | 12.4 ± 0.4 | 2.24 ± 0.08 | 3856 ± 55 | 1.36 ± 0.13 | 942 ± 3 |
| AEMS2 | 23.39 ± 0.25 | 13.3 ± 0.4 | 2.35 ± 0.08 | 4070 ± 58 | 1.64 ± 0.14 | 944 ± 2 |
| HSVI-BFS | 23.40 ± 0.25 | 13.0 ± 0.4 | 2.30 ± 0.08 | 3573 ± 52 | 1.69 ± 0.27 | 946 ± 3 |

Table 4: Comparison of different search heuristics on different instances of the *FieldVision-RockSample* environment.

### 4.4.1 Real-Time Performance of Online Search

In Table 4, we present 95% confidence intervals on the mean for our metrics of interest. We consider two instances of this environment, FVRS[5,5] (801 states, 5 actions, 32 observations) and FVRS[5,7] (3201 states, 5 actions, 128 observations). In both cases, we use the QMDP upper bound and Blind lower bound, under real-time constraints of 1 second per action.

**Return**  In terms of return, we do not observe any clear winner. BI-POMDP performs surpringly well in FVRS[5,5] but significantly worse than AEMS2 and HSVI-BFS in FVRS[5,7]. On the other hand, AEMS2 does significantly better than HSVI-BFS in FVRS[5,5] but both get very similar performances in FVRS[5,7]. Satia and Lave performs better in this environment than in RockSample. This is likely due to the fact that the transitions in belief space are no longer deterministic (as was the case with the *Move* actions in RockSample). In FVRS[5,5], we also observe that even when RTBSS is given 3 seconds per action to perform a two-step lookahead, its performance is worse than any of the heuristic search methods. This clearly shows that expanding all observations equally in the search is not a good strategy, as many of these observations can have negligible impact for the current decision.

**EBR and LBI**  In terms of error bound reduction and lower bound improvement, we observe that AEMS2 performs much better than HSVI-BFS in FVRS[5,5], but not significantly better in FVRS[5,7]. On the other hand, BI-POMDP obtains similar results to AEMS2 in FVRS[5,5] but does significantly worse in terms of EBR and LBI than in FVRS[5,7]. This suggests that AEMS2 is consistently effective at reducing the error, even in environments with large branching factors.

**Nodes Reused**  The percentage of belief nodes reused is much lower in FVRS due to the much higher branching factor. We observe that HSVI-BFS has the best reuse percentage in all environments, however not significantly higher than AEMS2. Both of these methods reuse significantly larger portion of the tree than the other methods. This confirms that these two methods are able to guide the search towards the most likely beliefs.

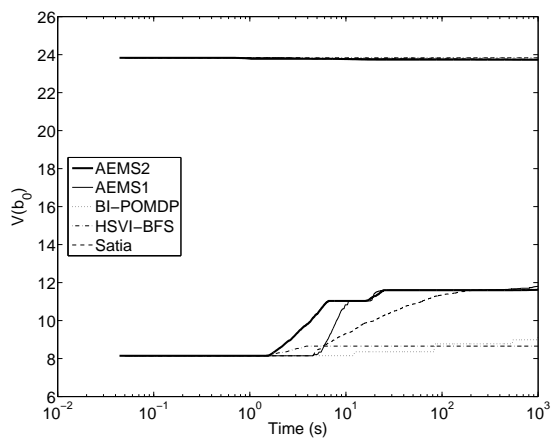 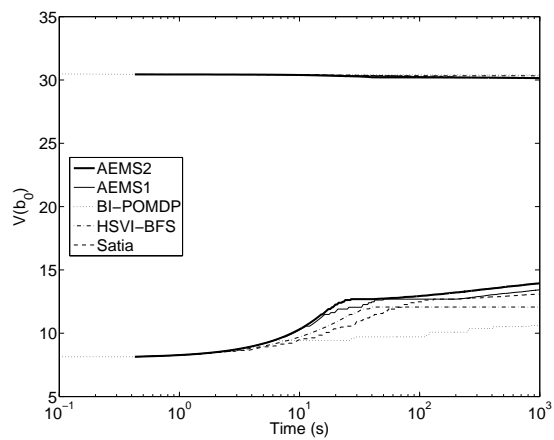

Figure 7: Evolution of the upper and lower bounds on *FieldVisionRockSample[5,5]*.

Figure 8: Evolution of the upper and lower bounds on *FieldVisionRockSample[5,7]*.

### 4.4.2 LONG-TERM ERROR REDUCTION OF ONLINE HEURISTIC SEARCH

Overall, while Table 4 confirms the consistent performance of HSVI-BFS and AEMS2, the difference with other heuristics is modest. Considering the complexity of this environment, this may be due to the fact that the algorithms do not have enough time to expand a significant number of nodes within 1 second. The long-term analysis of the bounds' evolution in Figures 7 and 8 confirms this. We observe in these figures that the lower bound converges slightly more rapidly with AEMS2 than with other heuristics. The AEMS1 heuristic also performs well in the long run on this problem, and seems to be the second best heuristic, while Satia and Lave is not far behind. On the other hand, the HSVI-BFS heuristic is far worse in this problem than in RockSample. This seems to be in part due to the fact that this heuristic takes more time to find the next node to expand than the others, and thus explores fewer belief states.

## 5. Discussion

The previous sections presented and evaluated several online POMDP algorithms. We now discuss important issues that arise when applying online methods in practice, and summarize some of their advantages and disadvantages. This should help researchers decide whether online algorithms are a good approach for solving a given problem.

## 5.1 Lower and Upper Bound Selection

Online algorithms can be combined with many valid lower and upper bounds. However, there are some properties that these bounds should satisfy for the online search to perform efficiently in practice. One of the desired properties is that the lower and upper bound functions should be monotone. The monotone property states that $\forall b : L(b) \leq \max_{a \in A} \left[ R_B(b,a) + \gamma \sum_{z \in Z} \Pr(z|b,a)L(\tau(b,a,z)) \right]$ for the lower bound and $\forall b : U(b) \geq \max_{a \in A} \left[ R_B(b,a) + \gamma \sum_{z \in Z} \Pr(z|b,a)U(\tau(b,a,z)) \right]$ for the upper bound. This property guarantees that when a certain fringe node is expanded, its lower bound is non-decreasing and its upper bound is non-increasing. This is sufficient to guarantee that the error bound $U_T(b) - L_T(b)$ on $b$ is non-increasing after the expansion of $b$, such that the error bound given by the algorithm on the value of the root belief state $b_c$, cannot be worse than the error bound defined by the initial bounds given. Note however that monotonicity is not necessary for AEMS to converge to an $\epsilon$-optimal solution, as shown in previous work (Ross et al., 2008); boundedness is sufficient.

## 5.2 Improving the Bounds over Time

As we mentioned in our survey of online algorithms, one drawback of many online approaches is that they do not store improvements made to the offline bounds during the online search, such that, if the same belief state is encountered again, the same computations need to be performed again, restarting from the same offline bounds. A trivial way to improve this is to maintain a large hashtable (or database) of belief states for which we have improved the lower and upper bounds in previous search, with their associated new bounds. There are however many drawbacks in doing this. First every time we want to evaluate the lower and upper bound of a fringe belief, a search through this hashtable needs to be performed to check if we have better bounds available. This may require significant time if the hashtable is large (e.g. millions of beliefs). Furthermore, experiments conducted with RTDP-Bel in large domains, such as RockSample[7,8], have shown that such a process usually runs out of memory (i.e. requires more than 1 GB) before good performance is achieved and requires several thousands episodes before performing well (Paquet, 2006).

The authors of RTBSS have also tried combining their search algorithm with RTDP-Bel such as to preserve the improvements made by the search (Paquet, 2006). While this combination usually performed better and learned faster than RTDP-Bel alone, it was found that in most domains, a few thousand episodes are still required before any improvement can be seen (in terms of return). Hence, such point updates of the offline bounds tend to be useful in large domains only if the task to accomplish is repeated a very large number of times.

A better strategy to improve the lower bound might be to save some time to perform $\alpha$-vector updates for some of the beliefs expanded during the search, such that the offline lower bound improves over time. Such updates have the advantage of improving the lower bound over the whole belief space, instead of at a single belief state. However this can be very time consuming, especially in large domains. Hence, if we need to act within very short time constraints, such an approach is infeasible. However if several seconds of planning time are available per action, then it might be advantageous to use some of this time to perform $\alpha$-vector updates, rather than use all the available time to search through the tree. A good idea here would be to perform $\alpha$-vector updates for a subset of the beliefs in the search tree, where the lower bound most improves.

## 5.3 Factored POMDP Representations

The efficiency of online algorithms relies heavily on the ability to quickly compute $\tau(b, a, z)$ and $\Pr(z|b, a)$, as these must be computed for evey belief state in the search tree. Using factored POMDP representations is an effective way to reduce the time complexity of computing these quantities. Since most environments with large state spaces are structured and can be described by sets of features, obtaining factored representation of complex systems should not be an issue in most cases. However, in domains with significant dependencies between state features, it may be useful to use algorithms proposed by Boyen and Koller (1998) and Poupart (2005) to find approximate factored representations where most features are independent, with minimal degradation in the solution quality. While the upper and lower bounds might not hold anymore if they are computed over the approximate factored representation, usually it may still yield good results in practice.

## 5.4 Handling Graph Structure

As we have mentioned before, the general tree search algorithm used by online algorithms will duplicate belief states whenever there are multiple paths leading to the same posterior belief from the current belief $b_c$. This greatly simplifies the complexity related to updating the values of ancestor nodes, and it also reduces the complexity related to finding the best fringe node to expand (using the technique in Section 3.4 which is only valid for trees). The disadvantage of using a tree structure is that inevitably, some computations will be redundant, as the algorithm will potentially expand the same subtree under every duplicate belief. To avoid this, we could use the $LAO^*$ algorithm proposed by Hansen and Zilberstein (2001) as an extension of $AO^*$ that can handle generic graph structure, including cyclic graphs. After each expansion, it runs a value (or policy) iteration algorithm until convergence among all ancestor nodes in order to update their values.

The heuristics we surveyed in Section 3.4 can be generalized to guide best-first search algorithms that handle graph structure, like $LAO^*$. The first thing to notice is that, in any graph, if a fringe node is reached by multiple paths, then its error contributes multiple times to the error on the value of $b_c$. Under this error contribution perspective, the heuristic value of such a fringe node should be the sum of its heuristic values over all paths reaching it. For instance, in the case of the AEMS heuristic, using the notation we have defined in Section 3.4, the global heuristic value of a given fringe node $b$, on the current belief state $b_c$ in any graph $G$, can be computed as follows:

$$H_G(b_c, b) = (U(b) - L(b)) \sum_{h \in \mathcal{H}_G(b_c, b)} \gamma^{d(h)} \Pr(h|b_c, \hat{\pi}_G). \tag{38}$$

Notice that for cyclic graphs, there can be infinitely many paths in $\mathcal{H}_G(b_c, b)$. In such case, we could use dynamic programming to estimate the heuristic value.

Because solving $H_G(b_c, b)$ for all fringe nodes $b$ in the graph $G$ will require a lot of time in practice, especially if there are many fringe nodes, we have not experimented with this method in Section 4. However, it would be practical to use this heuristic if we could find an alternative way to determine the best fringe node without computing $H_G(b_c, b)$ separately for each fringe node $b$ and performing an exhaustive search over all fringe nodes.

## 5.5 Online vs. Offline Time

One important aspect determining the efficiency and applicability of online algorithms is the amount of time available during the execution for planning. This of course is often task-dependent. For real-time problems like robot navigation, this amount of time may be very short, e.g. between 0.1 to 1 second per action. On the other hand for tasks like portfolio management, where acting every second is not necessary, several minutes could easily be taken to plan any stock buying/selling action. As we have seen from our experiments, the shorter the available online planning time, the greater the importance of having a good offline value function to start with. In such case, it is often necessary to reserve sufficient time to compute a good offline policy. As more and more planning time is available online, the influence of the offline value function becomes negligible, such that a very rough offline value function is sufficient to obtain good performance. The best trade-off between online and offline time often depends on how large the problem is. When the branching factor ($|A||Z|$) is large and/or computing successor belief states takes a long time, then more online time will be required to achieve a significant improvement over the offline value function. However, for small problems, an online time of 0.1 second per action may be sufficient to perform near-optimally even with a very rough offline value function.

## 5.6 Advantages and Disadvantages of Online Algorithms

We now discuss the advantages and disadvantages of online planning algorithms in general.

### 5.6.1 Advantages

- Most online algorithms can be combined with any offline solving algorithm, assuming it provides a lower bound or an upper bound on $V^*$, such as to improve the quality of the policy found offline.

- Online algorithms require very little offline computation before being executable in an environment, as they can perform well even using very loose bounds, which are quick to compute.

- Online methods can exploit the knowledge of the current belief to focus computation on the most relevant future beliefs for the current decision, such that they scale well to large action and observation spaces.

- Anytime online methods are applicable in real-time environments, as they can be stopped whenever planning time runs out, and still provide the best solution found so far.

### 5.6.2 Disadvantages

- The branching factor depends on the number of actions and observations. Thus if there are many observations and/or actions, it might be impossible to search deep enough, to provide significant improvement of the offline policy. In such cases, sampling methods designed to reduce the branching factor could be useful. While we cannot guarantee that the lower and upper bounds are still valid when sampling is used, we can guarantee that they are valid with high probability, given that enough samples are drawn.

- Most online algorithms do not store improvements made to the offline policy by the online search, and so the algorithm has to plan again with the same bounds each time the environment is restarted. If time is available, it could be advantageous to add $\alpha$-vector updates for some belief states explored in the tree, so that the offline bounds improve with time.

## 6. Conclusion

POMDPs provide a rich and elegant framework for planning in stochastic partially observable domains, however their time complexity has been a major issue preventing their application to complex real-world systems. This paper thoroughly surveys the various existing online algorithms and the key techniques and approximations used to solve POMDPs more efficiently. We empirically compare these online approaches in several POMDP domains under different metrics: average discounted return, average error bound reduction and average lower bound improvement, and using different lower and upper bounds: PBVI, Blind, FIB and QMDP.

From the empirical results, we observe that some of the heuristic search methods, namely AEMS2 and HSVI-BFS, obtain very good performances, even in domains with large branching factors and large state spaces. These two methods are very similar and perform well because they orient the search towards nodes that can improve the current approximate value function as quickly as possible; i.e. the belief nodes that have largest error and are most likely to be reached in the future by "promising" actions. However, in environments with large branching factors, we may only have time to expand a few nodes at each turn. Hence, it would be interesting to develop further approximations to reduce the branching factor in such cases.

In conclusion, we believe that online approaches have an important role to play in improving the scalability of POMDP solution methods. A good example is the succesful applications of the RTBSS algorithm to the RobocupRescue simulation by Paquet et al. (2005). This environment is very challenging as the state space is orders of magnitude beyond the scope of current algorithms. Offline algorithms remain very important to obtain tight lower and upper bounds on the value function. The interesting question is not whether online or offline approaches are better, but how we can improve both kinds of approaches, such that their synergy can be exploited to solve complex real-world problems.

## Acknowledgments

This research was supported by the Natural Sciences and Engineering Council of Canada and the Fonds Québécois de la Recherche sur la Nature et les Technologies. We would also like to thank the anonymous reviewers for their helpful comments and suggestions.

# References

Astrom, K. J. (1965). Optimal control of Markov decision processes with incomplete state estimation. *Journal of Mathematical Analysis and Applications*, *10*, 174–205.

Barto, A. G., Bradtke, S. J., & Singhe, S. P. (1995). Learning to act using real-time dynamic programming. *Artificial Intelligence*, *72*(1), 81–138.

Bellman, R. (1957). *Dynamic Programming*. Princeton University Press, Princeton, NJ, USA.

Bertsekas, D. P., & Castanon, D. A. (1999). Rollout algorithms for stochastic scheduling problems. *Journal of Heuristics*, *5*(1), 89–108.

Boyen, X., & Koller, D. (1998). Tractable inference for complex stochastic processes. In *In Proceedings of the Fourteenth Conference on Uncertainty in Artificial Intelligence (UAI-98)*, pp. 33–42.

Braziunas, D., & Boutilier, C. (2004). Stochastic local search for POMDP controllers. In *The Nineteenth National Conference on Artificial Intelligence (AAAI-04)*, pp. 690–696.

Cassandra, A., Littman, M. L., & Zhang, N. L. (1997). Incremental pruning: a simple, fast, exact method for partially observable Markov decision processes. In *Proceedings of the Thirteenth Conference on Uncertainty in Artificial Intelligence (UAI-97)*, pp. 54–61.

Chang, H. S., Givan, R., & Chong, E. K. P. (2004). Parallel rollout for online solution of partially observable Markov decision processes. *Discrete Event Dynamic Systems*, *14*(3), 309–341.

Geffner, H., & Bonet, B. (1998). Solving large POMDPs using real time dynamic programming. In *Proceedings of the Fall AAAI symposium on POMDPs*, pp. 61–68.

Hansen, E. A. (1998). Solving POMDPs by searching in policy space. In *Fourteenth Conference on Uncertainty in Artificial Intelligence (UAI-98)*, pp. 211–219.

Hansen, E. A., & Zilberstein, S. (2001). LAO * : A heuristic search algorithm that finds solutions with loops. *Artificial Intelligence*, *129*(1-2), 35–62.

Hauskrecht, M. (2000). Value-function approximations for partially observable Markov decision processes. *Journal of Artificial Intelligence Research*, *13*, 33–94.

Kaelbling, L. P., Littman, M. L., & Cassandra, A. R. (1998). Planning and acting in partially observable stochastic domains. *Artificial Intelligence*, *101*, 99–134.

Kearns, M. J., Mansour, Y., & Ng, A. Y. (1999). A sparse sampling algorithm for near-optimal planning in large markov decision processes. In *Proceedings of the Sixteenth International Joint Conference on Artificial Intelligence (IJCAI-99)*, pp. 1324–1331.

Koenig, S. (2001). Agent-centered search. *AI Magazine*, *22*(4), 109–131.

Littman, M. L. (1996). *Algorithms for sequential decision making*. Ph.D. thesis, Brown University.

Littman, M. L., Cassandra, A. R., & Kaelbling, L. P. (1995). Learning policies for partially observable environments: scaling up. In *Proceedings of the 12th International Conference on Machine Learning (ICML-95)*, pp. 362–370.

Lovejoy, W. S. (1991). Computationally feasible bounds for POMDPs. *Operations Research*, *39*(1), 162–175.

Madani, O., Hanks, S., & Condon, A. (1999). On the undecidability of probabilistic planning and infinite-horizon partially observable Markov decision problems. In *Proceedings of the Sixteenth National Conference on Artificial Intelligence. (AAAI-99)*, pp. 541–548.

McAllester, D., & Singh, S. (1999). Approximate Planning for Factored POMDPs using Belief State Simplification. In *Proceedings of the 15th Annual Conference on Uncertainty in Artificial Intelligence (UAI-99)*, pp. 409–416.

Monahan, G. E. (1982). A survey of partially observable Markov decision processes: theory, models and algorithms. *Management Science*, *28*(1), 1–16.

Nilsson, N. (1980). *Principles of Artificial Intelligence*. Tioga Publishing.

Papadimitriou, C., & Tsitsiklis, J. N. (1987). The complexity of Markov decision processes. *Mathematics of Operations Research*, *12*(3), 441–450.

Paquet, S. (2006). *Distributed Decision-Making and Task Coordination in Dynamic, Uncertain and Real-Time Multiagent Environments*. Ph.D. thesis, Laval University.

Paquet, S., Chaib-draa, B., & Ross, S. (2006). Hybrid POMDP algorithms. In *Proceedings of The Workshop on Multi-Agent Sequential Decision Making in Uncertain Domains (MSDM-06)*, pp. 133–147.

Paquet, S., Tobin, L., & Chaib-draa, B. (2005). An online POMDP algorithm for complex multiagent environments. In *Proceedings of The fourth International Joint Conference on Autonomous Agents and Multi Agent Systems (AAMAS-05)*, pp. 970–977.

Pineau, J., Gordon, G., & Thrun, S. (2003). Point-based value iteration: an anytime algorithm for POMDPs. In *Proceedings of the International Joint Conference on Artificial Intelligence (IJCAI-03)*, pp. 1025–1032.

Pineau, J., Gordon, G., & Thrun, S. (2006). Anytime point-based approximations for large POMDPs. *Journal of Artificial Intelligence Research*, *27*, 335–380.

Pineau, J. (2004). *Tractable planning under uncertainty: exploiting structure*. Ph.D. thesis, Carnegie Mellon University.

Poupart, P. (2005). *Exploiting structure to efficiently solve large scale partially observable Markov decision processes*. Ph.D. thesis, University of Toronto.

Poupart, P., & Boutilier, C. (2003). Bounded finite state controllers. In *Advances in Neural Information Processing Systems 16 (NIPS)*.

Puterman, M. L. (1994). *Markov Decision Processes: Discrete Stochastic Dynamic Programming*. John Wiley & Sons, Inc.

Ross, S., & Chaib-draa, B. (2007). Aems: An anytime online search algorithm for approximate policy refinement in large POMDPs. In *Proceedings of the 20th International Joint Conference on Artificial Intelligence (IJCAI-07)*, pp. 2592–2598.

Ross, S., Pineau, J., & Chaib-draa, B. (2008). Theoretical analysis of heuristic search methods for online POMDPs. In *Advances in Neural Information Processing Systems 20 (NIPS)*.

Satia, J. K., & Lave, R. E. (1973). Markovian decision processes with probabilistic observation of states. *Management Science, 20*(1), 1–13.

Shani, G., Brafman, R., & Shimony, S. (2005). Adaptation for changing stochastic environments through online POMDP policy learning. In *Proceedings of the Workshop on Reinforcement Learning in Non-Stationary Environments, ECML 2005*, pp. 61–70.

Smallwood, R. D., & Sondik, E. J. (1973). The optimal control of partially observable Markov processes over a finite horizon. *Operations Research, 21*(5), 1071–1088.

Smith, T., & Simmons, R. (2004). Heuristic search value iteration for POMDPs. In *Proceedings of the 20th Conference on Uncertainty in Artificial Intelligence (UAI-04)*, pp. 520–527.

Smith, T., & Simmons, R. (2005). Point-based POMDP algorithms: improved analysis and implementation. In *Proceedings of the 21th Conference on Uncertainty in Artificial Intelligence (UAI-05)*, pp. 542–547.

Sondik, E. J. (1971). *The optimal control of partially observable Markov processes*. Ph.D. thesis, Stanford University.

Sondik, E. J. (1978). The optimal control of partially observable Markov processes over the infinite horizon: Discounted costs. *Operations Research, 26*(2), 282–304.

Spaan, M. T. J., & Vlassis, N. (2004). A point-based POMDP algorithm for robot planning. In *In Proceedings of the IEEE International Conference on Robotics and Automation (ICRA-04)*, pp. 2399–2404.

Spaan, M. T. J., & Vlassis, N. (2005). Perseus: randomized point-based value iteration for POMDPs. *Journal of Artificial Intelligence Research, 24*, 195–220.

Vlassis, N., & Spaan, M. T. J. (2004). A fast point-based algorithm for POMDPs. In *Benelearn 2004: Proceedings of the Annual Machine Learning Conference of Belgium and the Netherlands*, pp. 170–176.

Washington, R. (1997). BI-POMDP: bounded, incremental partially observable Markov model planning. In *Proceedings of the 4th European Conference on Planning*, pp. 440–451.

Zhang, N. L., & Zhang, W. (2001). Speeding up the convergence of value iteration in partially observable Markov decision processes. *Journal of Artificial Intelligence Research, 14*, 29–51.